\documentclass{article}

\PassOptionsToPackage{numbers, compress}{natbib}

\usepackage[preprint]{neurips/neurips_2026}

\usepackage[utf8]{inputenc} 
\usepackage[T1]{fontenc}    
\usepackage{hyperref}       
\usepackage{url}            
\usepackage{booktabs}       
\usepackage{amsfonts}       
\usepackage{nicefrac}       
\usepackage{microtype}      
\usepackage{xcolor}         

\title{Task‑Adaptive Embedding Refinement \\ via Test-time LLM Guidance}

\author{Ariel Gera, Shir Ashury-Tahan, Gal Bloch, Ohad Eytan, Assaf Toledo \\
IBM Research \\
\texttt{\{ariel.gera1,assaf.toledo\}@ibm.com}}

\usepackage{algorithm}
\usepackage{algorithmic}

\usepackage{subcaption}

\usepackage{epigraph}
\usepackage{enumitem}
\usepackage{multirow}
\usepackage[most]{tcolorbox}

\newcommand{\ALGCOMMENT}[1]{\hfill\textcolor{gray}{$\triangleright$ #1}}

\newcommand\realscholarlong[0]{\textsc{RealScholarQuery}}
\newcommand\realscholar[0]{\textsc{RealScholar}}
\newcommand\kpm[0]{\textsc{ArgKP-21}}
\newcommand\banking[0]{\textsc{Banking77}}
\newcommand\clinc[0]{\textsc{CLINC150}}
\newcommand\followir[0]{\textsc{FollowIR}}
\newcommand\nfcorpus[0]{\textsc{NFCorpus}}

\newcommand\qwensmall[0]{Qwen3-Embedding-0.6B}
\newcommand\qwenlarge[0]{Qwen3-Embedding-8B}
\newcommand\nemotron[0]{Llama-Embed-Nemotron-8B}
\newcommand\efivemistral[0]{E5-Mistral-7B}
\newcommand\linq[0]{Linq-Embed-Mistral}

\newcommand\mistralsmall[0]{Mistral-Small-3.2-24B}
\newcommand\deepseek[0]{DeepSeek-V3.2}
\newcommand\qwenllm[0]{Qwen3.5-397B-A17B-FP8}
\newcommand\qwenllmshort[0]{Qwen3.5}
\newcommand\gptfour[0]{GPT-4.1}

\usepackage[capitalize,noabbrev]{cleveref}

\begin{document}

\maketitle

\begin{abstract}
We explore the effectiveness of an \textit{LLM-guided query refinement} paradigm for extending the usability of embedding models to challenging zero-shot search and classification tasks. Our approach refines the embedding representation of a user query using feedback from a generative LLM on a small set of documents, enabling embeddings to adapt in real time to the target task. We conduct extensive experiments with state-of-the-art text embedding models across a diverse set of challenging search and classification benchmarks. Empirical results indicate that LLM-guided query refinement yields consistent gains across all models and datasets, with relative improvements of up to +25\% in literature search, intent detection, key-point matching, and nuanced query-instruction following. The refined queries improve ranking quality and induce clearer binary separation across the corpus, enabling the embedding space to better reflect the nuanced, task-specific constraints of each ad-hoc user query. Importantly, this expands the range of practical settings in which embedding models can be effectively deployed, making them a compelling alternative when costly LLM pipelines are not viable at corpus-scale. We release our experimental code for reproducibility\footnote{\url{https://github.com/IBM/task-aware-embedding-refinement}}.
\end{abstract}

\section{Introduction}
Embedding models have long been explored for a broad range of applications over large corpora
\citep{reimers-gurevych-2019-sentence,thakur2021beir,muennighoff-etal-2023-mteb,enevoldsen2025mmteb}. A prominent example is the paradigm of Retrieval-Augmented Generation (RAG; \citealp{lewis2020rag,gao2024ragsurvey}), which typically relies on embedding models to compute dense semantic representations over an entire corpus, enabling efficient online ranking via query-document similarity scores. 

However, these models are still a long way from the flexibility and instruction-following capabilities that generative LLMs offer out-of-the-box. In particular, zero-shot classification with a general-purpose embedding model, in response to an ad-hoc user input, remains a major challenge.

\begin{figure}
    \centering
    \includegraphics[clip,trim=0 338 0 3,width=0.95\linewidth]{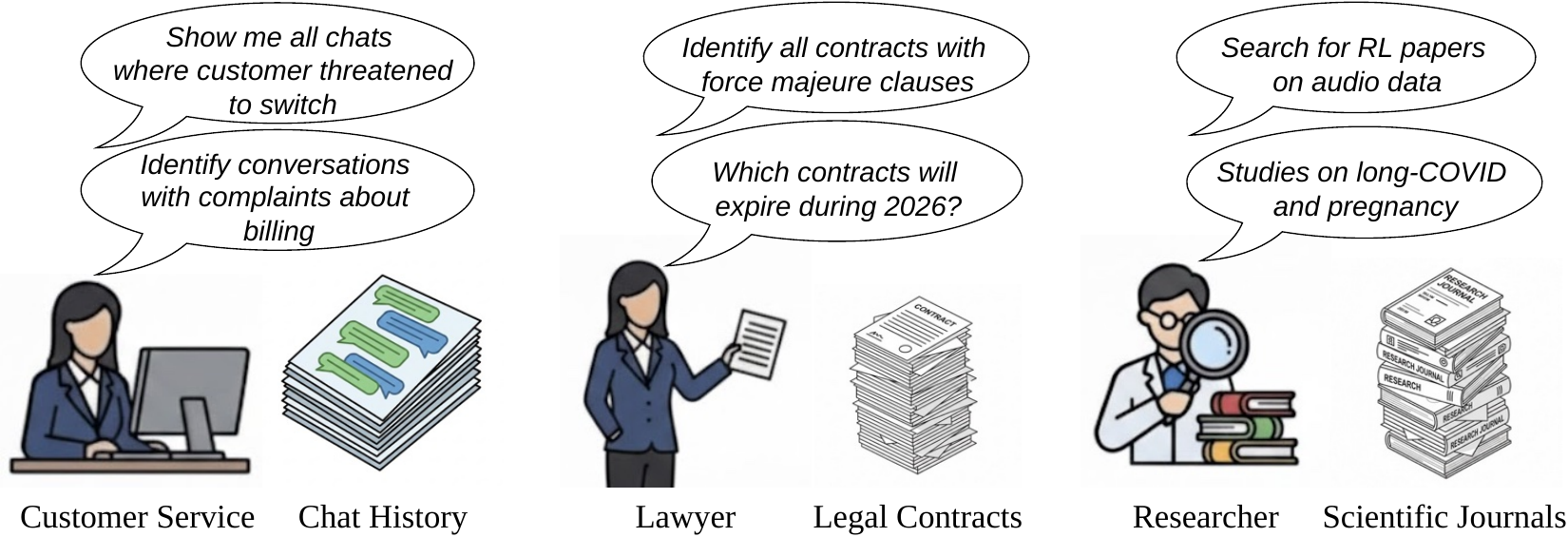}
    \caption{Examples of full-corpus separation tasks over different collections of documents.} 
    \label{fig:tasks_examples}
    \vspace{-.5cm}
\end{figure}

Thus, although embedding models provide substantially better trade-offs than generative LLMs in terms of runtime cost and latency -- rendering them the only practical solution for large-scale, real-world deployments -- their effectiveness for zero-shot classification remains severely limited. Figure~\ref{fig:tasks_examples} illustrates some of these use-cases.

For instance, consider a business use case of an analyst that wishes to review different patterns within a massive corpus of conversations -- for example, quantifying and analyzing customer support chats where the customer threatened to switch providers. This requires classifying the entire document corpus according to an ad-hoc binary criterion.
Similarly, for a literature search system, often the user aims to get \textit{all} relevant documents, effectively separating the full corpus into positive and negative instances. The efficiency constraints of such classification and search tasks therefore make a compelling case for tackling them with embedding models -- provided that they provide sufficient representational quality.

Recent work has demonstrated the potential of guided query refinement \citep{gangireddy2025rerankerrelevancefeedback,uzan2025guidedqueryrefinementmultimodal}, which leverages feedback from an encoder or reranker to optimize representations and improve ranking performance in traditional document retrieval tasks. However, the broader applicability of this paradigm remains largely unexplored.

In this work we examine the value of the query refinement paradigm for enhancing the expressiveness of embedding models. We utilize feedback from a general-purpose LLM at test time, harnessing it to extend their applicability to challenging search and classification scenarios.

In extensive experiments over multiple leading embedding models and a wide variety of classification and search tasks across different domains, we show that limited test-time LLM feedback can transform the embedding-based ranking of the document space, adapting it to a specific user query. This is reflected in consistent gains in Mean Average Precision (MAP) for the refined query compared with the original query across all evaluated tasks: academic literature search \citep{gusenbauer2020academic} ($+16.9\%$), intent detection \citep{larson2019clinc150} ($+9.4\%$), key-point matching \citep{friedman-etal-2021-overview} ($+15\%$), and nuanced query instructions \citep{weller-etal-2025-followir} ($+7.4\%$). Averaged over all models and tasks, the refined query yields a MAP improvement of $12\%$.


By combining the zero-shot flexibility of generative LLMs with the efficiency of embedding-based pipelines, our hybrid approach enables embedding models to move beyond retrieval and offline fine-tuned classification settings into the realm of zero-shot full-corpus classification across diverse domains. In doing so, it achieves a favorable balance between task accuracy and scalability to large-scale corpora.

\section{Related Work}

\paragraph{General-Purpose Embedding Models}

Embedding models have been 
studied across a wide range of applications, including retrieval and search as well as diverse classification tasks \citep{cer-etal-2018-universal, enevoldsen2025mmteb, gao-etal-2021-simcse, karpukhin2020densepassageretrievalopendomain, colbert, muennighoff-etal-2023-mteb, reimers-gurevych-2019-sentence,thakur2021beir}. The diversity of use cases has driven the development of increasingly capable ``universal'' embedding models, which seek to learn high-quality semantic representations that generalize effectively across tasks and domains. Recent models \citep{qwen3embedding, lee2025geminiembeddinggeneralizableembeddings, babakhin2025llamaembednemotron8buniversaltextembedding} have demonstrated impressive gains on benchmarks such as MTEB \cite{muennighoff-etal-2023-mteb}, which measure embedding quality across a range of target tasks.
The advances in general-purpose dense retriever models have partly been driven by progress in generative LLMs -- in particular, by using decoder-only LLMs as a backbone and starting point for training the embedding models, and also for generating synthetic training data \citep{zuccon2025r2llms}.

Multiple works have imbued embedders with instruction-following capabilities, starting with early efforts like TART \citep{asai-etal-2023-task} and InstructOR \citep{su-etal-2023-one} that incorporated basic task information into the query. Later works, utilizing LLM-generated synthetic training data, have incorporated this approach into large-scale instruction-tuning pipelines \citep{wang-etal-2024-improving-text,lee2025nvembed,qwen3embedding,babakhin2025llamaembednemotron8buniversaltextembedding}. Beyond task 
instructions, \citet{li2025making} have explored an option of supplying the embedder with few-shot demonstrations with each query.


Even when used with dedicated task instructions, embedding models still exhibit difficulties with the long tail of diverse and specialized information retrieval tasks \citep{sun-etal-2024-mair}. Specifically, they struggle with complex and nuanced query descriptions \citep{weller-etal-2025-followir}, as well as retrieval that requires deep reasoning \citep{su2025bright}.


\paragraph{Test-time Query Refinement}
Several retrieval works have explored the refinement of query representations during test time, often termed as a form of ``pseudo-relevance feedback'' (PRF). The feedback consists of scores from a separate retriever or cross-encoder reranker for some top-ranked documents, which are used to refine the embedding representation for the query by the target retriever model \citep{yu2021improving,sung-etal-2023-optimizing,gangireddy2025rerankerrelevancefeedback,uzan2025guidedqueryrefinementmultimodal}. Here we largely follow the refinement formulations from ReFIT \citep{gangireddy2025rerankerrelevancefeedback} and GQR \citep{uzan2025guidedqueryrefinementmultimodal}, but use generative LLMs to provide the feedback scores. Existing works focus on a traditional retrieval setup (e.g., for QA); in contrast, here we broaden the scope to explore the adaptability of embedding models for specialized and nuanced zero-shot classification and search tasks. For a detailed comparison with ReFIT and GQR, see Appendix~\ref{app:prior_work}.




\section{Method}
We begin by describing the general problem formulation (\S\ref{ssec:problem}). We then present the details of our approach (\S\ref{ssec:optimization}).

\subsection{Problem Formulation} \label{ssec:problem}
Our focus in this work is on tasks that entail a binary separation problem over a large-scale corpus.

Formally, given a query $q$ and a collection of documents $D=\{d_i\}_{i=1}^{N}$, the goal is to identify the subset of positive documents $D_{pos} \subseteq D$ that are a match for $q$.

Note that this formulation covers both search and classification scenarios (Fig. \ref{fig:tasks_examples}). In search, the task is often framed as asking which documents are \textit{relevant} for the query;\footnote{For simplicity, we treat relevance as a binary property.} in classification, we are seeking documents that meet a certain positive relationship with the query. For example, these could be pairs where $d$ \textit{discusses} the topic $q$, where $d$ \textit{entails} $q$, where $d$ is \textit{evidence} for $q$, and so on.

We denote a ranking of $D$ as an ordered list $\pi_D=[d_{\pi(i)}]_{i=1}^{N}$. In addition, we denote the index of a document $d \in D$ in the ranking $\pi_D$ by $IND_{\pi_D}(d)$. An optimal ranking of $D$ with respect to a query $q$ is one where every positive document appears before any negative document:
\[
\forall{d^{+}\in D_{pos}, d^{-}\in D_{neg}}: \  IND_{\pi_D}(d^{+}) < IND_{\pi_D}(d^{-})
\]
where $D_{neg} := D \setminus D_{pos}$.

In our setup, we take a query $q$ and use it to create an initial, typically sub-optimal, ranking of the document collection -- $\pi_{q, D}$. Our task is to improve upon the initial ranking and derive a high-quality ranking that can better serve as a basis for separating the positive documents from the negative ones, with respect to the given $q$.\footnote{Here we do not address the question of how to set the threshold beyond which documents are assumed to be negative; we discuss this below in \S\ref{ssec:metrics}.}

\begin{algorithm*}[t]
  \caption{\emph{LLM-Guided Test-Time Query Optimization}}
  \label{alg:test_time_optimization}
  \begin{algorithmic}[1]
    \REQUIRE Query $q$, embedding model $m_e$, teacher model $m_t$, iterations $T$, step size $\alpha$, feedback set size $K$
    
    \STATE $\mathcal{D}_{K} \gets \pi_{q, D}[:K]$ \ALGCOMMENT{Ranking of $D$ wrt. $q$ using $m_e$, and taking top-$K$}

    \STATE $\mathbf{p}_{m_t}(q) \gets \operatorname{softmax}([s_{m_t}(q,d)$ for $d \in \mathcal{D}_K])$ \ALGCOMMENT{Teacher model distribution over $\mathcal{D}_K$}

    \STATE $z^{(0)} \gets m_e(q)$ \ALGCOMMENT{Initialization of optimized $q$ embedding}

    \FOR{$t=0$ \textbf{to} $T-1$}
        \STATE $\mathbf{p}_{m_e}(z^{(t)}) \gets \operatorname{softmax}([s_{m_e}(z^{(t)},m_e(d))$ for $d \in \mathcal{D}_K]$) \ALGCOMMENT{Embedding model distribution over $\mathcal{D}_K(q)$}

        \STATE $\mathcal{L}(t) \gets \operatorname{KL}\!\big(\mathbf{p}_{m_t}(q) \,\|\, \mathbf{p}_{m_e}(z^{(t)})\big)$ \ALGCOMMENT{Compute the KL divergence loss}

        \STATE $z^{(t+1)} \gets z^{(t)} - \alpha\, \nabla_{z^{(t)}} \mathcal{L}(t)$  \ALGCOMMENT{Gradient step on the query representation}
    \ENDFOR

    \STATE \textbf{return} $\pi_{z^{(T)},D}$ \ALGCOMMENT{Ranking of $D$ wrt. the optimized $q$ using $m_e$}

  \end{algorithmic}
\end{algorithm*}

\subsection{Test-Time Query Optimization Algorithm} \label{ssec:optimization}
Our approach leverages an embedding model to generate an initial ranking over the document corpus, and subsequently improves this ranking by incorporating feedback from a more capable -- though computationally expensive -- teacher model. The full procedure is described below, and summarized in Algorithm \ref{alg:test_time_optimization}.

Formally, an embedding model $m_e$ encodes representations $m_e(q)$ and $m_e(d)$ 
for each query and document in a shared vector space.  
The induced ranking $\pi_{q, D}$ is obtained by applying the similarity function $S_{m_e}$ which the model was optimized for (e.g., cosine similarity) to pairs of $m_e(q)$ and $m_e(d)$, for all $d \in D$, and ordering the documents in descending order. 

Following \citet{gangireddy2025rerankerrelevancefeedback} and \citet{uzan2025guidedqueryrefinementmultimodal}, we utilize feedback scores to refine the query representation given by $m_e$. 
In our approach, these feedback scores are given by a \textit{teacher model} $m_t$, which is a generative LLM. $m_t$ jointly attends over the query and document, to directly output scores $s_{m_t}(q,d)$ for the query-document pairs. 

Let $\mathcal{D}_K=\big(d^{(1)},\dots,d^{(K)}\big)$
denote the top-$K$ documents from $\pi_{q, D}$.
$m_t$ produces a vector $\mathbf{s}_{m_t}(q)$ of teacher scores
\[ \mathbf{s}_{m_t}(q) =
\bigl(s_{m_t}(q,d^{(1)}), \dots, s_{m_t}(q,d^{(K)})\bigr)
\]



We now use $\mathbf{s}_{m_t}(q)$ to perform a \textbf{test-time optimization} of the query embedding. 
We denote the initial embedding as \(z^{(0)} = m_e(q)\), updating it at each optimization step $t$, \(z^{(t)}\) for $T$ steps.
$\mathbf{s}_{m_e}(z^{(t)})$ is the vector of the similarity scores over the top-$K$ documents at step $t$.

We minimize
\[
\mathcal{L}^{(t)}=\mathrm{KL}\!\big(p_{m_t}(q)\,\|\,p_{m_e}(z^{(t)})\big)
\]
where \(\mathrm{KL}\) is the Kullback–Leibler divergence, and where $p_{m_t}$ and $p_{m_e}$ are the softmax over $\mathbf{s}_{m_t}(q)$ and $\mathbf{s}_{m_e}(z^{(t)})$.

We apply a gradient step on the query representation with step size $\alpha$,
\[
z^{(t+1)}=z^{(t)}-\alpha\,\nabla_z \mathcal{L}\!\big(z^{(t)}\big)
\]

After $T$ steps, we compute updated scores for $m_e$ over the full corpus, based on the refined representation $z^{(T)}$:

\[
s_{m_e}^{(T)}(q,d)=s_{m_e}\!\big(z^{(T)}, d\big) \quad \text{for } d\in{D}.
\]

We use a single fixed configuration of 
parameters for all datasets, models and settings (\S\ref{ssec:opt_implementation}).

Thus, we apply test-time optimization that uses a feedback signal from $m_t$ to modify the query representation $m_e(q)$; we then use the refined query to calculate new scores and induce a new ranking. Note that the procedure is performed online for $q$, affecting only the representation of this specific query -- without modifying the $m_e$ model weights or the representations of corpus documents.

\section{Experiments}
To investigate the adaptability of embedding models for zero-shot binary separation, we conduct experiments over varied search and classification datasets (\S\ref{ssec:tasks}), using state-of-the-art embedding models. In the experiments we compare the naive ranking performance to performance when using our query optimization approach (\S\ref{ssec:metrics}).

\subsection{Tasks and Datasets} \label{ssec:tasks}

Below we describe some target tasks and datasets we explore. We find these to be illustrative of the range of scenarios where our hybrid modeling approach can prove beneficial. However, we stress that they are merely examples from a broader task space of relevant problems.

\paragraph{Literature Search}
A common real-world search task is academic literature search \citep{gusenbauer2020academic}, where the goal is to find a set of papers that meet a unique set of relevance criteria (e.g., ``Provide me with studies that proposed hierarchical neural models to capture spatiotemporal features in sign
videos''). This task is particularly challenging: it requires specialized knowledge, understanding of domain-specific terminology, and the space of possible user queries is arbitrarily large; moreover, users typically require very high recall -- consider, for instance, an author of a survey paper who does not wish to overlook relevant papers. At the same time, such queries are expected to run on massive corpora and return a result within a reasonable time frame. This makes this a compelling use case for combining fast dense retrieval with a feedback signal from a knowledgeable LLM teacher. Here we rely on \realscholarlong{} \citep{he-etal-2025-pasa}: a dataset of real-world queries over a massive corpus of computer science papers from arXiv. Document relevance was comprehensively annotated by experts, with multiple matching documents for each query.

\paragraph{Intent Analytics} 
Many businesses maintain corpora of conversations, e.g., customer support interactions, and require the ability to analyze and quantify specific subtypes of conversations from such large-scale data. In the NLP literature, a prominent example is the task of intent detection \citep{liu2019review}, where the goal is to identify why the customer contacted the organization and what they aim to achieve. Typically, the task is formulated as classifying a novel utterance into predefined intents; however, another possibility is an online analytics setting -- where users query the full corpus of utterances, for candidate intent categories they consider post-hoc.
This and similar tasks are common in real-world use cases; for instance, business analysts may wish to investigate different subsets of conversations where the user had a specific need, yet this requires separating the corpus to focus on the ad-hoc criterion or subset of interest. As a representative for this kind of binary detection task, we take \clinc{} \citep{larson2019clinc150}, which classifies customer service requests from $10$ domains (such as Banking, Travel, and Work) into $150$ user intents. In our setting, we treat each intent category as a query, and test the ability to separate utterances that express the intent from those that do not. LLM feedback 
helps the embedding model
align the brief query text with the latent structure of a target intent.  

\paragraph{Key-Point Matching}
Key-point analysis \citep{bar-haim-etal-2020-arguments} is an approach for obtaining a high-level quantitative summary of a large collection of opinions. It constitutes of mapping free-text inputs into a set of high-level ``key points'', and then \textit{quantifying} the proportion of inputs corresponding to each point. This is used, for example, to gain insights from open-ended responses to large-scale surveys. As noted by \citet{eden-etal-2023-welcome}, it is difficult to scale such solutions to large-scale collections, where the major computational bottleneck is the \textit{matching} between the key points and individual input texts in the corpus. Here we experiment with \kpm{}, key-point matching data from the 2021 KPA shared task \citep{friedman-etal-2021-overview}; each query corresponds to a key point, and the goal is accurate pairwise matching.

\paragraph{Nuanced Query Instructions}
Real-world users often formulate queries with strict constraints that go beyond simple topical matching, requiring the retrieval system to handle explicit exclusions and fine-grained scoping. This is particularly challenging for dense retrievers, as they must distinguish between documents that are semantically similar and those that actually satisfy the user's logical conditions (e.g., ``retrieve accomplishments of the Hubble telescope but exclude details of repairs"). Here we utilize \textsc{FollowIR} \citep{weller-etal-2025-followir}, a benchmark derived from rigorous TREC relevance narratives. Unlike standard short queries, these instructions feature complex logic and negation, serving as a stress test for the ability to suppress "easy" semantic matches that violate specific user instructions.

\paragraph{Additional Datasets}
In Appendix \ref{app:additional} we present results for additional tasks: \banking{} \citep{casanueva-etal-2020-efficient}, which classifies customer service requests from the banking domain into $77$ fine-grained user intents;
and \textsc{NFCorpus} \citep{boteva2016full}, in which lay user queries are used to retrieve relevant medical literature.

Dataset details are provided in Appendix \ref{app:datasets}.

\subsection{Evaluation} \label{ssec:metrics}
In our main experiments we measure the quality of the induced rankings, comparing results with the base query embedding to results after the query refinement procedure described in \S\ref{ssec:optimization}. 

Our primary metric for task separation quality is the mean average precision (MAP) across the dataset queries. 
As explained in \S\ref{ssec:problem}, our focus here is on the quality of the induced ranking. While the ultimate goal is often to obtain the final separation into $D_{pos}$ and $D_{neg}$, the strategy for setting the appropriate score threshold depends on the specifics of the use-case. For instance, in a literature search it may be that the end users themselves will determine the threshold, whereas for intent detection it would be desirable to set the threshold automatically for each query. We note that one could apply a simple heuristic to choose the threshold, for instance by progressively sampling additional documents, sending them to the LLM, and using some rule-based stopping criterion or a threshold-tuning approach (see e.g., \citealp{chausson2025insight,  vannooten2025sizedoesfitall}). Importantly, threshold tuning is complementary to the problem of obtaining a high‑quality ranking, and we leave a systematic exploration of the interaction between these two aspects to future work.

In popular benchmarks like MTEB \citep{muennighoff-etal-2023-mteb}, the quality of embedding models for classification tasks is measured by training a classifier (e.g., logistic regression) over embedding representations, using train set gold labels; In contrast, here we do not assume the existence of any ground-truth labels for training, and focus on the zero-shot ranking quality for an unseen class (i.e., query).

\subsection{Implementation Details} \label{ssec:opt_implementation}

\paragraph{Optimization}
For the optimization procedure described in \S\ref{ssec:optimization}, we use the Adam \citep{kingma2015adam} optimizer. Based on the hyperparameter analysis given in \citet{uzan2025guidedqueryrefinementmultimodal}, in all of our experiments we set the learning rate $\alpha=10^{-4}$ and the number of steps $T=100$. We note that the latency of the optimization procedure for each query is under a second, rendering it a practical test-time solution.

In choosing the documents $\mathcal{D}_K(q)$ that are sent for LLM feedback, we opt for a naive approach of simply taking the top-$K$ documents (with $K=20$) according to the original embedding model scores ($s_{m_e}^{(0)}(q,d)$). Intuitively, this increases the likelihood of encountering some positive pairs in $\mathcal{D}_K(q)$, which may make the feedback signal more informative. We leave to future work the question of how to choose the optimal document collection for feedback.

\begin{figure*}[t]
    \centering

\subfloat{\small{(a)}
\includegraphics[width=0.44\textwidth]{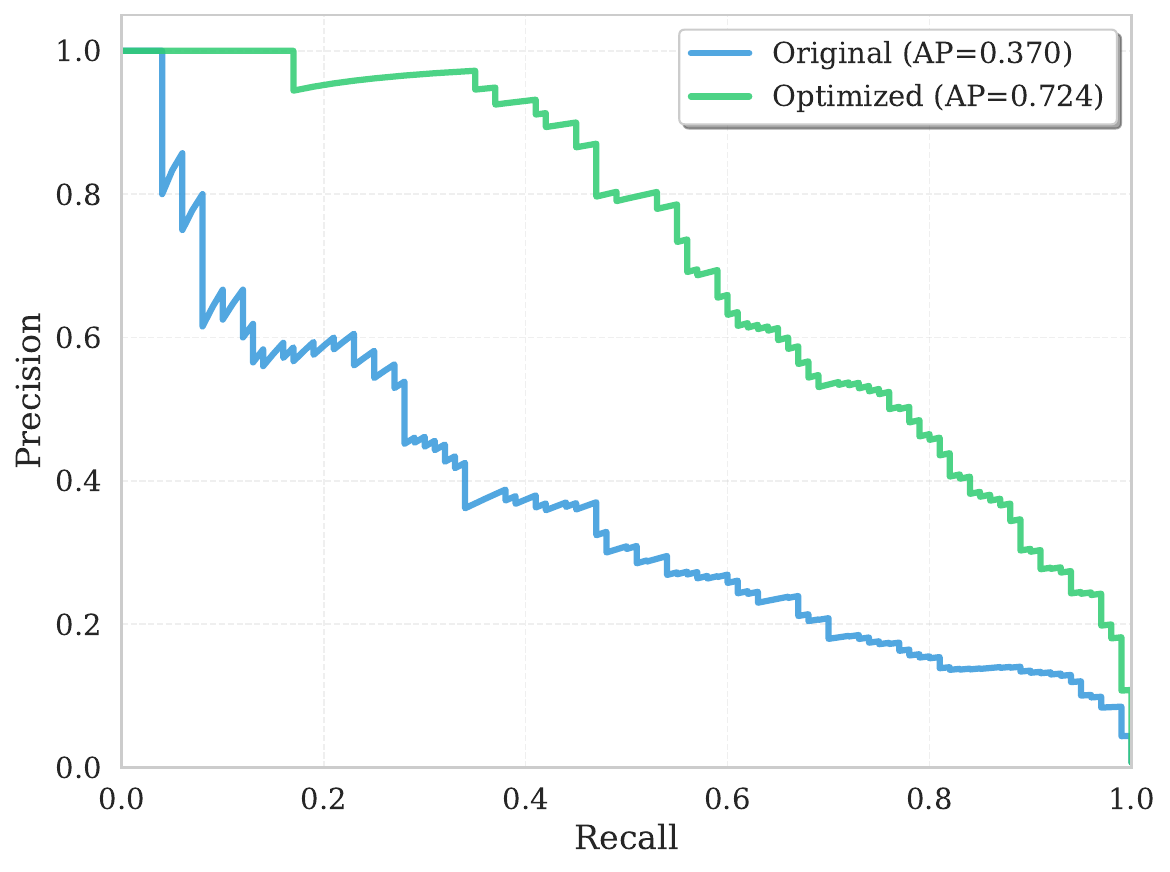} 
}
\subfloat{\small{(b)}
\includegraphics[width=0.45\textwidth]{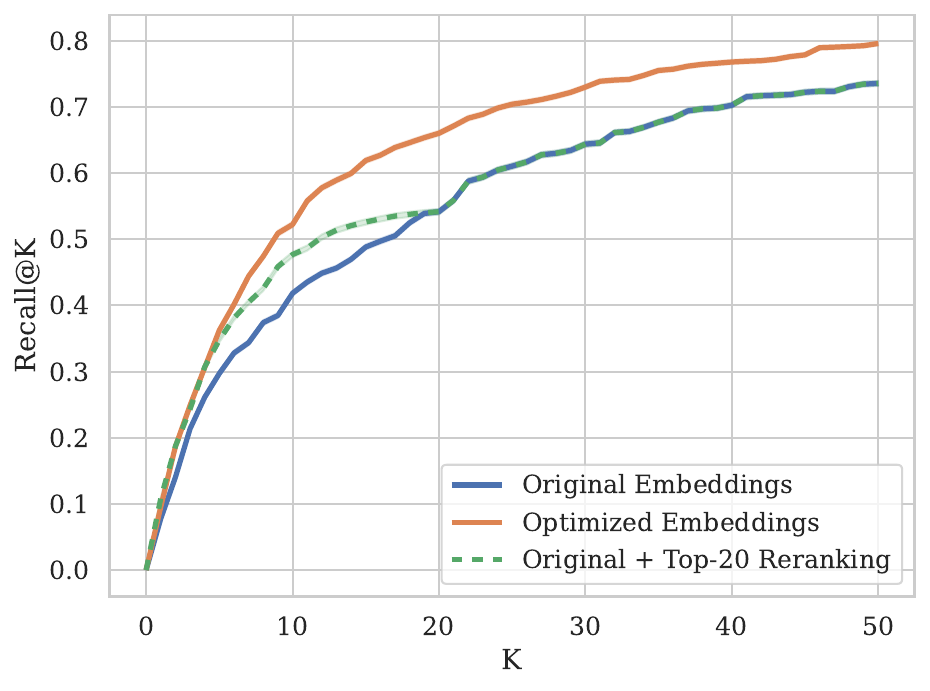} 
}
\caption{Example results of LLM-guided test-time query optimization.
(a) Precision-recall curve for an individual intent query from \clinc{} (with \textit{\qwensmall{}} as the embedding model), comparing results before and after the query refinement procedure. The plot legend denotes the corresponding average precision (AP) scores.
(b) Recall@K (aggregated across queries) for \textit{\efivemistral{}} on \realscholarlong{}. Query refinement is based on LLM feedback scores for the top-$20$ documents. The dashed line depicts results when using the original query embeddings to calculate the similarity scores, but reranking just the top-$20$ documents based on the feedback.
}
\label{fig:pr_examples}
\end{figure*}

\paragraph{Models} \label{ssec:models}
We examine a diverse set of leading embedding models: \qwensmall{}, \qwenlarge{} \citep{qwen3embedding}, \nemotron{} \citep{babakhin2025llamaembednemotron8buniversaltextembedding}, \linq{} \citep{choi2024linq} and \efivemistral{} \citep{wang-etal-2024-improving-text}.
In our main results the LLM feedback scores are provided by \mistralsmall{}-Instruct-2506; we test some additional LLMs in Appendix~\ref{app:additional}.

\paragraph{Inference Details}
To obtain feedback scores for a specific document, we send a query-document pair with a brief instruction that asks the LLM to judge whether this pair is a match. Scores are calculated based on the token logprobs of the generated response; refer to Appendix~\ref{app:scores} for details. 

\label{p:inference}
For the query-document similarity scores $s_{m_e}(q,d)$, we use the cosine similarity between the embeddings of $q$ and $d$.
As recommended in documentation of the various embedding models, we embed the query text as part of a query template that also includes a short task instruction:
{\small\verb|Instruct: {instruction}\nQuery: {query}|} \\We explore the effect of the embedding instruction in \S\ref{app:embed_instructions}.

\section{Results} \label{sec:results}


We compare ranking quality induced by the original query embedding representations to the ranking obtained after test-time query refinement (\S\ref{ssec:optimization}) based on LLM feedback.

Figure~\ref{fig:pr_examples}a depicts a precision-recall curve for an example query.
We see that the refined query representation induces a more favorable ranking, with a better ability to separate positive and negative documents. The improved precision-recall dynamics translate to better average precision (AP) scores.



Our procedure entails obtaining high-quality predictions from the teacher $m_t$ for documents $\mathcal{D}_K(q)$. Thus, we also measure performance of the original query embedding, but where the $m_t$ feedback scores are used to \textit{rerank} the specific documents included in $\mathcal{D}_K(q)$, i.e., without modifying the other similarity scores. 
In Figure~\ref{fig:pr_examples}b we show the Recall@K measurements for the ranking induced by the embedding similarity scores. Naturally, the effect of using LLM feedback directly for reranking (dashed line in the figure) is limited only to the top-K documents on which feedback was obtained; in contrast, we see that utilizing this feedback for query refinement increases recall beyond the initial top-K, improving the overall ranking of documents. 

Table~\ref{tab:main_results} presents the aggregated results across queries -- the mean average precision (MAP) score for different datasets and embedding models, with \textit{\mistralsmall} as the teacher $m_t$. As can be seen, the query refinement approach provides a substantial benefit across tasks and across embedding models. These improvements are statistically significant (see Appendix~\ref{app:additional}). We also see that query refinement provides a consistent benefit on top of the reranking-only setup, with the exception of \textit{\linq{}} embeddings. 

\begin{table*}
\centering
\caption{Mean average precision (MAP) results across queries. For each embedding model, we compare the ranking induced by the \textit{original} query embeddings to the one using \textit{optimized} query embeddings after LLM-guided query refinement (\S\ref{ssec:optimization}).} \label{tab:main_results}

\resizebox{\textwidth}{!}{%
\begin{tabular}{lllllll}
\toprule
 &  & \realscholar{} & \clinc{} & \kpm{} & \followir{} & Average\\
Embedding Model & Setting &  &  &  &  \\
\midrule
\multirow[t]{3}{*}{\qwensmall{}} & Original & .54 & .59 & .67 & .41 & .55 \\
 & Rerank-only & .62 {\scriptsize (+15.5\%)} & .60 {\scriptsize (+1.3\%)} & .73 {\scriptsize (+8.0\%)} & .45 {\scriptsize (+8.2\%)} & .60 {\scriptsize (+8.1\%)} \\
 & Optimized & .65 {\scriptsize (+20.7\%)} & .74 {\scriptsize (+25.5\%)} & .79 {\scriptsize (+17.2\%)} & .46 {\scriptsize (+11.3\%)} & .66 {\scriptsize (+19.2\%)} \\
\cline{1-7}
\multirow[t]{3}{*}{\qwenlarge{}} & Original & .61 & .86 & .77 & .45 & .67 \\
 & Rerank-only & .68 {\scriptsize (+12.5\%)} & .87 {\scriptsize (+0.1\%)} & .78 {\scriptsize (+2.0\%)} & .47 {\scriptsize (+5.5\%)} & .70 {\scriptsize (+4.3\%)} \\
 & Optimized & .70 {\scriptsize (+15.0\%)} & .88 {\scriptsize (+2.1\%)} & .83 {\scriptsize (+8.1\%)} & .49 {\scriptsize (+8.7\%)} & .72 {\scriptsize (+7.8\%)} \\
\cline{1-7}
\multirow[t]{3}{*}{\nemotron{}} & Original & .54 & .80 & .59 & .45 & .59 \\
 & Rerank-only & .65 {\scriptsize (+20.4\%)} & .80 {\scriptsize (+0.5\%)} & .67 {\scriptsize (+13.8\%)} & .46 {\scriptsize (+4.0\%)} & .65 {\scriptsize (+9.0\%)} \\
 & Optimized & .67 {\scriptsize (+23.0\%)} & .85 {\scriptsize (+6.5\%)} & .74 {\scriptsize (+25.7\%)} & .47 {\scriptsize (+5.1\%)} & .68 {\scriptsize (+14.7\%)} \\
\cline{1-7}
\multirow[t]{3}{*}{\linq{}} & Original & .51 & .76 & .67 & .43 & .59 \\
 & Rerank-only & .59 {\scriptsize (+16.8\%)} & .76 {\scriptsize (+0.6\%)} & .73 {\scriptsize (+8.1\%)} & .46 {\scriptsize (+7.0\%)} & .63 {\scriptsize (+7.4\%)} \\
 & Optimized & .51 {\scriptsize (+1.2\%)} & .76 {\scriptsize (+0.2\%)} & .68 {\scriptsize (+0.6\%)} & .43 {\scriptsize (+0.5\%)} & .59 {\scriptsize (+0.6\%)} \\
\cline{1-7}
\multirow[t]{3}{*}{\efivemistral{}} & Original & .48 & .69 & .63 & .41 & .55 \\
 & Rerank-only & .55 {\scriptsize (+15.1\%)} & .70 {\scriptsize (+0.9\%)} & .69 {\scriptsize (+10.4\%)} & .44 {\scriptsize (+7.3\%)} & .60 {\scriptsize (+7.8\%)} \\
 & Optimized & .60 {\scriptsize (+24.5\%)} & .78 {\scriptsize (+12.8\%)} & .77 {\scriptsize (+23.3\%)} & .46 {\scriptsize (+11.4\%)} & .65 {\scriptsize (+18.1\%)} \\
\cline{1-7}
\bottomrule
\end{tabular}
}
\end{table*}

\paragraph{Additional LLM Teachers}

In addition to \textit{\mistralsmall}, we test $3$ other LLM teacher models: \textit{\deepseek{}}, \textit{\qwenllmshort{}} and \textit{\gptfour{}}. We observe consistent performance across LLM teachers (Appendix~\ref{app:additional}).

\paragraph{Comparison with HyDE}

A popular approach for improving zero-shot retrieval performance is HyDE \citep{gao-etal-2023-hyde}, which refines the \textit{text} of the user query, by replacing it with an LLM-generated ``hypothetical'' document. Thus, we experiment with applying HyDE to the original queries. We find that unlike our query refinement approach, HyDE has a negative effect on performance; at the same time, our method works well also when applied on top of HyDE-style query texts (Appendix~\ref{app:hyde}),

\paragraph{Latency} Assuming offline indexing of the corpus documents, the online latency of running a user query consists of obtaining an \textit{initial query embedding}, obtaining an \textit{initial ranking} (i.e., calculating the similarity scores $s_{m_e}(q,d)$), and running the additional \textit{test-time query refinement} procedure. 
As shown in Table~\ref{tab:latency}, the query refinement steps entail a modest overhead of under $100$ms, plus the latency of obtaining the $K=20$ LLM feedback scores.


\begin{table*}
\centering
\caption{Per-query latencies in wall-clock time, on a single A100 80GB GPU. The measurements below were conducted with \textit{\qwenlarge{}} as the embedding model, over \clinc{}.} \label{tab:latency}


\begin{tabular}{llll}
\toprule
Online step & Mean & Median & Std \\
\midrule
Query embedding & $86.32$ms & $83.75$ms & $5.79$ms \\
Initial ranking & $1.42$ms & $1.44$ms &  $0.05$ms \\
Query optimization (excluding LLM calls) & $79.20$ms & $76.47$ms & $4.17$ms 
\\
\bottomrule
\end{tabular}

\end{table*}

\begin{figure}[h]
    \centering
\includegraphics[width=0.45\linewidth]{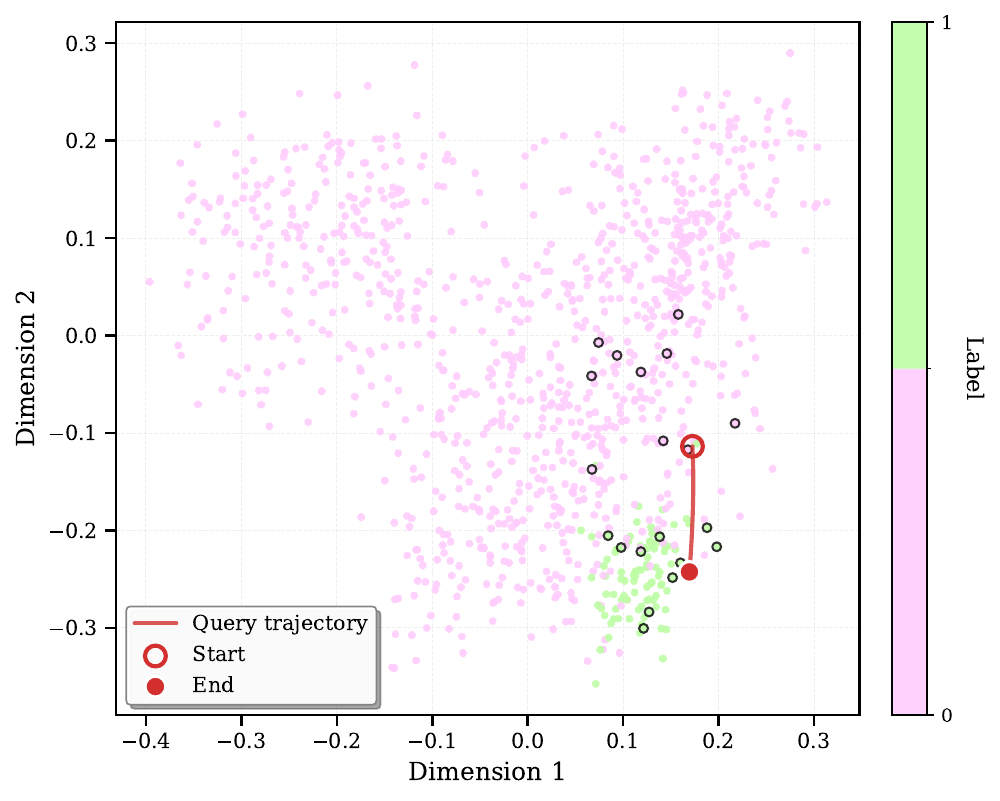}
    \caption{2D PCA projection for a query from \clinc{}. The plot depicts the trajectory of the query embedding during the $T=100$ gradient steps of test-time optimization.
    Each dot marks a document, and colors denote the gold label of the query-document pairs. Circled dots are the documents in $\mathcal{D}_K(q)$, which received LLM feedback scores and thus guided the optimization.}
    \label{fig:pca_example}

\end{figure}

\section{Analysis}

\paragraph{Query Trajectories} \label{ssec:refine_2d_analysis}

We have shown that refining the query embedding 
improves the ability of the induced ranking to capture a target binary separation.
Aiming to better understand the underlying mechanism, we use dimensionality reduction to visualize the changes in the query embedding.

Figure \ref{fig:pca_example} depicts a PCA 2D projection of the embedding space for a single query from \clinc{}. The plot illustrates the trajectory of the query representation as it undergoes optimization -- presented in the context of the space of the corpus document representations, as well as the ground-truth labels. In this example it appears that the LLM feedback scores for a set of positive and negative documents guided the query representation towards a more discriminative region of the embedding space -- one that better separates positive examples from negative ones. However, as we demonstrate in Appendix~\ref{app:2d}, the query trajectory patterns are more complex, and vary significantly between queries.



\paragraph{Impact of K} \label{ssec:k_analysis}

\begin{figure}
    \centering
    \includegraphics[width=.6\linewidth]{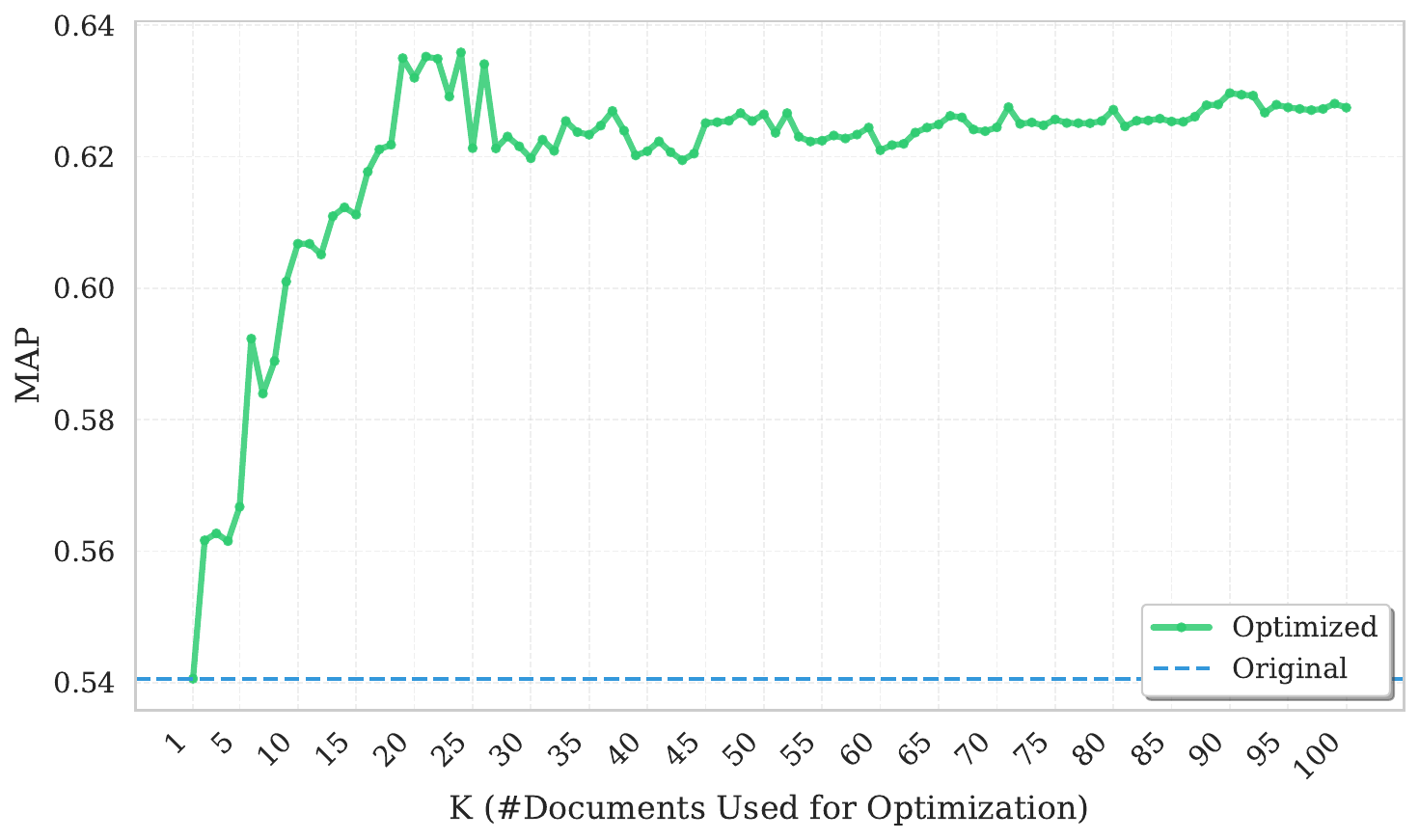}
    \caption{MAP scores as a function of the number of documents $K$ used for optimization. The plot depicts results for \textit{\qwensmall{}} on \realscholarlong{}.}
    \label{fig:k_plot}
    \vspace{-.5cm}
\end{figure}

Since the optimization depends on feedback from the top-K retrieved documents, we would expect that increasing the number of ranked documents will enhance the optimization quality. It is reasonable to assume a law of diminishing returns, where the marginal benefit of each additional document decreases as the set size grows. Both of these trends are illustrated in \cref{fig:k_plot}. However, determining the optimal value for K remains an open question for future work.

\paragraph{Impact of Feedback Quality}
The method naturally depends on the feedback scores given by the LLM teacher. While we have shown in \S\ref{sec:results} that different teachers have quite similar performance overall, we also directly test the connection between the per-query correctness of the LLM feedback and the outcome of optimization for that query. This analysis reveals modest but statistically significant positive correlations between feedback quality and performance gain (Appendix Table~\ref{tab:perf_correlations}).

\paragraph{Effect of Embedding Instructions} \label{ssec:instructions}
Given our focus on tackling varied tasks with embedding models, we also perform an analysis of the effect of the task instruction given in the query (see \S\ref{p:inference}), as well as the query instruction template used. We find that some models (in particular the smaller \textit{\qwensmall{}}) are more sensitive to these choices; we also find that for specialized tasks, using the models without a task instruction can be detrimental. Refer to Appendix~\ref{app:embed_instructions} for details.

\section{Discussion}

In this work, we explored the paradigm of LLM-guided query refinement, as a way to extend the applicability of embedding models. We have shown that this test-time optimization achieves substantial gains across varied tasks and domains. Importantly, this expands the space of possibilities for a broad range of real-world search and classification needs over massive corpora, where costly LLM-centric pipelines are not a viable solution.

More broadly, this demonstrates the power of a hybrid modeling approach, one that utilizes and combines different kinds of models according to their strengths and weaknesses -- relying on the potential of embedding models for scalability and efficiency, but also harnessing the nuanced understanding and zero-shot versatility of LLMs.

Our results show that even a relatively small set of targeted feedback scores at test time can help adapt a query embedding, relocating it in the vector space for a better separation of the data.
Here we opted for a simple approach to select this small collection 
used for feedback -- taking the $K$ top-ranked documents.
However, we see potential for exploring more sophisticated methods for constructing the feedback set. This evokes the established field of active learning \citep{cohn1996active,zhang-etal-2022-survey,ein-dor-etal-2020-active}, which asks what 
are the best
instances to send for human labeling, in order to support model training; the applicable question here is which instances should be sent for LLM scoring, to be most useful for test-time feedback.

While our approach offers a highly efficient and low-latency solution for task-adaptive retrieval, it is subject to several limitations. First, the method’s efficacy is intrinsically linked to the quality of the teacher model’s feedback; consequently, any systematic biases or conceptual errors within the teacher may be encoded into the refined query representations. Second, although our top-$K$ selection strategy ensures rapid online optimization, it may overlook more informative documents that lie outside the initial retrieval set, particularly in scenarios involving extreme class imbalance. 

Future work could explore the role of teacher model selection, trade-offs between teacher capability and inference cost, and whether smaller, domain-specialized teachers could provide efficient yet high-quality feedback. Moreover, the efficiency of the feedback step can be increased even further, by replacing pointwise relevance scoring with list-wise or set-wise scoring strategies \citep{Zhuang_2024}. 


Finally, this paradigm can easily be extended beyond text to additional modalities, for instance for use in image classification and visual detection tasks. This could enable similar test-time adaptation benefits across an even broader range of real-world use cases.


\section{Conclusion}
We introduced LLM‑guided query refinement as a test‑time mechanism for extending the applicability of embedding models. We show that even with a small amount of targeted feedback, this approach yields substantial performance gains across diverse datasets, 
enabling effective search and classification over massive corpora without relying on prohibitively expensive LLM‑centric pipelines. Our findings underscore the value of hybrid modeling strategies that leverage the scalability and efficiency of embedding models while selectively incorporating nuanced reasoning capabilities of LLMs.

\bibliographystyle{plainnat}
\bibliography{custom}

\begin{thebibliography}{45}
\providecommand{\natexlab}[1]{#1}
\providecommand{\url}[1]{\texttt{#1}}
\expandafter\ifx\csname urlstyle\endcsname\relax
  \providecommand{\doi}[1]{doi: #1}\else
  \providecommand{\doi}{doi: \begingroup \urlstyle{rm}\Url}\fi

\bibitem[Asai et~al.(2023)Asai, Schick, Lewis, Chen, Izacard, Riedel, Hajishirzi, and Yih]{asai-etal-2023-task}
Akari Asai, Timo Schick, Patrick Lewis, Xilun Chen, Gautier Izacard, Sebastian Riedel, Hannaneh Hajishirzi, and Wen-tau Yih.
\newblock Task-aware retrieval with instructions.
\newblock In Anna Rogers, Jordan Boyd-Graber, and Naoaki Okazaki, editors, \emph{Findings of the Association for Computational Linguistics: ACL 2023}, pages 3650--3675, Toronto, Canada, July 2023. Association for Computational Linguistics.
\newblock \doi{10.18653/v1/2023.findings-acl.225}.
\newblock URL \url{https://aclanthology.org/2023.findings-acl.225/}.

\bibitem[Babakhin et~al.(2025)Babakhin, Osmulski, Ak, Moreira, Xu, Schifferer, Liu, and Oldridge]{babakhin2025llamaembednemotron8buniversaltextembedding}
Yauhen Babakhin, Radek Osmulski, Ronay Ak, Gabriel Moreira, Mengyao Xu, Benedikt Schifferer, Bo~Liu, and Even Oldridge.
\newblock Llama-{E}mbed-{N}emotron-8{B}: A universal text embedding model for multilingual and cross-lingual tasks.
\newblock \emph{arXiv:2511.07025}, 2025.
\newblock URL \url{https://arxiv.org/abs/2511.07025}.

\bibitem[Bar-Haim et~al.(2020)Bar-Haim, Eden, Friedman, Kantor, Lahav, and Slonim]{bar-haim-etal-2020-arguments}
Roy Bar-Haim, Lilach Eden, Roni Friedman, Yoav Kantor, Dan Lahav, and Noam Slonim.
\newblock From arguments to key points: {T}owards automatic argument summarization.
\newblock In Dan Jurafsky, Joyce Chai, Natalie Schluter, and Joel Tetreault, editors, \emph{Proceedings of the 58th Annual Meeting of the Association for Computational Linguistics}, pages 4029--4039, Online, July 2020. Association for Computational Linguistics.
\newblock \doi{10.18653/v1/2020.acl-main.371}.
\newblock URL \url{https://aclanthology.org/2020.acl-main.371/}.

\bibitem[Boteva et~al.(2016)Boteva, Gholipour, Sokolov, and Riezler]{boteva2016full}
Vera Boteva, Demian Gholipour, Artem Sokolov, and Stefan Riezler.
\newblock A full-text learning to rank dataset for medical information retrieval.
\newblock In \emph{European Conference on Information Retrieval}, pages 716--722. Springer, 2016.
\newblock URL \url{https://doi.org/10.1007/978-3-319-30671-1_58}.

\bibitem[Casanueva et~al.(2020)Casanueva, Tem{\v{c}}inas, Gerz, Henderson, and Vuli{\'c}]{casanueva-etal-2020-efficient}
I{\~n}igo Casanueva, Tadas Tem{\v{c}}inas, Daniela Gerz, Matthew Henderson, and Ivan Vuli{\'c}.
\newblock Efficient intent detection with dual sentence encoders.
\newblock In Tsung-Hsien Wen, Asli Celikyilmaz, Zhou Yu, Alexandros Papangelis, Mihail Eric, Anuj Kumar, I{\~n}igo Casanueva, and Rushin Shah, editors, \emph{Proceedings of the 2nd Workshop on Natural Language Processing for Conversational AI}, pages 38--45, Online, July 2020. Association for Computational Linguistics.
\newblock \doi{10.18653/v1/2020.nlp4convai-1.5}.
\newblock URL \url{https://aclanthology.org/2020.nlp4convai-1.5/}.

\bibitem[Cer et~al.(2018)Cer, Yang, Kong, Hua, Limtiaco, St.~John, Constant, Guajardo-Cespedes, Yuan, Tar, Strope, and Kurzweil]{cer-etal-2018-universal}
Daniel Cer, Yinfei Yang, Sheng-yi Kong, Nan Hua, Nicole Limtiaco, Rhomni St.~John, Noah Constant, Mario Guajardo-Cespedes, Steve Yuan, Chris Tar, Brian Strope, and Ray Kurzweil.
\newblock Universal sentence encoder for {E}nglish.
\newblock In Eduardo Blanco and Wei Lu, editors, \emph{Proceedings of the 2018 Conference on Empirical Methods in Natural Language Processing: System Demonstrations}, pages 169--174, Brussels, Belgium, November 2018. Association for Computational Linguistics.
\newblock \doi{10.18653/v1/D18-2029}.
\newblock URL \url{https://aclanthology.org/D18-2029/}.

\bibitem[Chausson et~al.(2025)Chausson, Fourcade, Harding, Ross, and Renard]{chausson2025insight}
Sandrine Chausson, Marion Fourcade, David~J Harding, Bj{\"o}rn Ross, and Gr{\'e}gory Renard.
\newblock The insight-inference loop: Efficient text classification via natural language inference and threshold-tuning.
\newblock \emph{Sociological Methods \& Research}, page 00491241251326819, 2025.
\newblock URL \url{https://doi.org/10.1177/00491241251326819}.

\bibitem[Choi et~al.(2024)Choi, Kim, Lee, Kwon, Gu, Kim, Cho, and Sohn]{choi2024linq}
Chanyeol Choi, Junseong Kim, Seolhwa Lee, Jihoon Kwon, Sangmo Gu, Yejin Kim, Minkyung Cho, and Jy-yong Sohn.
\newblock Linq-{E}mbed-{M}istral technical report.
\newblock \emph{arXiv:2412.03223}, 2024.
\newblock URL \url{https://arxiv.org/abs/2412.03223}.

\bibitem[Cohn et~al.(1996)Cohn, Ghahramani, and Jordan]{cohn1996active}
DA~Cohn, Z~Ghahramani, and MI~Jordan.
\newblock Active learning with statistical models.
\newblock \emph{The Journal of Artificial Intelligence Research}, 4:\penalty0 129, 1996.

\bibitem[Eden et~al.(2023)Eden, Kantor, Orbach, Katz, Slonim, and Bar-Haim]{eden-etal-2023-welcome}
Lilach Eden, Yoav Kantor, Matan Orbach, Yoav Katz, Noam Slonim, and Roy Bar-Haim.
\newblock Welcome to the real world: Efficient, incremental and scalable key point analysis.
\newblock In Mingxuan Wang and Imed Zitouni, editors, \emph{Proceedings of the 2023 Conference on Empirical Methods in Natural Language Processing: Industry Track}, pages 483--491, Singapore, December 2023. Association for Computational Linguistics.
\newblock \doi{10.18653/v1/2023.emnlp-industry.46}.
\newblock URL \url{https://aclanthology.org/2023.emnlp-industry.46/}.

\bibitem[Ein-Dor et~al.(2020)Ein-Dor, Halfon, Gera, Shnarch, Dankin, Choshen, Danilevsky, Aharonov, Katz, and Slonim]{ein-dor-etal-2020-active}
Liat Ein-Dor, Alon Halfon, Ariel Gera, Eyal Shnarch, Lena Dankin, Leshem Choshen, Marina Danilevsky, Ranit Aharonov, Yoav Katz, and Noam Slonim.
\newblock {A}ctive {L}earning for {BERT}: {A}n {E}mpirical {S}tudy.
\newblock In Bonnie Webber, Trevor Cohn, Yulan He, and Yang Liu, editors, \emph{Proceedings of the 2020 Conference on Empirical Methods in Natural Language Processing (EMNLP)}, pages 7949--7962, Online, November 2020. Association for Computational Linguistics.
\newblock \doi{10.18653/v1/2020.emnlp-main.638}.
\newblock URL \url{https://aclanthology.org/2020.emnlp-main.638/}.

\bibitem[Enevoldsen et~al.(2025)Enevoldsen, Chung, Kerboua, Kardos, Mathur, Stap, Gala, Siblini, Krzemi{\'n}ski, Winata, Sturua, Utpala, Ciancone, Schaeffer, Misra, Dhakal, Rystr{\o}m, Solomatin, {\c{C}}a{\u{g}}atan, Kundu, Bernstorff, Xiao, Sukhlecha, Pahwa, Po{\'s}wiata, GV, Ashraf, Auras, Pl{\"u}ster, Harries, Magne, Mohr, Zhu, Gisserot-Boukhlef, Aarsen, Kostkan, Wojtasik, Lee, Suppa, Zhang, Rocca, Hamdy, Michail, Yang, Faysse, Vatolin, Thakur, Dey, Vasani, Chitale, Tedeschi, Tai, Snegirev, Hendriksen, G{\"u}nther, Xia, Shi, L{\`u}, Clive, K, Anna, Wehrli, Tikhonova, Panchal, Abramov, Ostendorff, Liu, Clematide, Miranda, Fenogenova, Song, Safi, Li, Borghini, Cassano, Hansen, Hooker, Xiao, Adlakha, Weller, Reddy, and Muennighoff]{enevoldsen2025mmteb}
Kenneth Enevoldsen, Isaac Chung, Imene Kerboua, M{\'a}rton Kardos, Ashwin Mathur, David Stap, Jay Gala, Wissam Siblini, Dominik Krzemi{\'n}ski, Genta~Indra Winata, Saba Sturua, Saiteja Utpala, Mathieu Ciancone, Marion Schaeffer, Diganta Misra, Shreeya Dhakal, Jonathan Rystr{\o}m, Roman Solomatin, {\"O}mer~Veysel {\c{C}}a{\u{g}}atan, Akash Kundu, Martin Bernstorff, Shitao Xiao, Akshita Sukhlecha, Bhavish Pahwa, Rafa{\l} Po{\'s}wiata, Kranthi~Kiran GV, Shawon Ashraf, Daniel Auras, Bj{\"o}rn Pl{\"u}ster, Jan~Philipp Harries, Lo{\"\i}c Magne, Isabelle Mohr, Dawei Zhu, Hippolyte Gisserot-Boukhlef, Tom Aarsen, Jan Kostkan, Konrad Wojtasik, Taemin Lee, Marek Suppa, Crystina Zhang, Roberta Rocca, Mohammed Hamdy, Andrianos Michail, John Yang, Manuel Faysse, Aleksei Vatolin, Nandan Thakur, Manan Dey, Dipam Vasani, Pranjal~A Chitale, Simone Tedeschi, Nguyen Tai, Artem Snegirev, Mariya Hendriksen, Michael G{\"u}nther, Mengzhou Xia, Weijia Shi, Xing~Han L{\`u}, Jordan Clive, Gayatri K, Maksimova Anna, Silvan Wehrli, Maria
  Tikhonova, Henil~Shalin Panchal, Aleksandr Abramov, Malte Ostendorff, Zheng Liu, Simon Clematide, Lester James~Validad Miranda, Alena Fenogenova, Guangyu Song, Ruqiya~Bin Safi, Wen-Ding Li, Alessia Borghini, Federico Cassano, Lasse Hansen, Sara Hooker, Chenghao Xiao, Vaibhav Adlakha, Orion Weller, Siva Reddy, and Niklas Muennighoff.
\newblock {MMTEB}: Massive multilingual text embedding benchmark.
\newblock In \emph{The Thirteenth International Conference on Learning Representations}, 2025.
\newblock URL \url{https://openreview.net/forum?id=zl3pfz4VCV}.

\bibitem[Friedman et~al.(2021)Friedman, Dankin, Hou, Aharonov, Katz, and Slonim]{friedman-etal-2021-overview}
Roni Friedman, Lena Dankin, Yufang Hou, Ranit Aharonov, Yoav Katz, and Noam Slonim.
\newblock Overview of the 2021 key point analysis shared task.
\newblock In Khalid Al-Khatib, Yufang Hou, and Manfred Stede, editors, \emph{Proceedings of the 8th Workshop on Argument Mining}, pages 154--164, Punta Cana, Dominican Republic, November 2021. Association for Computational Linguistics.
\newblock \doi{10.18653/v1/2021.argmining-1.16}.
\newblock URL \url{https://aclanthology.org/2021.argmining-1.16/}.

\bibitem[Gangi~Reddy et~al.(2025)Gangi~Reddy, Dasigi, Sultan, Cohan, Sil, Ji, and Hajishirzi]{gangireddy2025rerankerrelevancefeedback}
Revanth Gangi~Reddy, Pradeep Dasigi, Md~Arafat Sultan, Arman Cohan, Avirup Sil, Heng Ji, and Hannaneh Hajishirzi.
\newblock A large-scale study of reranker relevance feedback at inference.
\newblock In \emph{Proceedings of the 48th International ACM SIGIR Conference on Research and Development in Information Retrieval}, SIGIR '25, page 3010–3014, New York, NY, USA, 2025. Association for Computing Machinery.
\newblock ISBN 9798400715921.
\newblock URL \url{https://doi.org/10.1145/3726302.3730160}.

\bibitem[Gao et~al.(2023)Gao, Ma, Lin, and Callan]{gao-etal-2023-hyde}
Luyu Gao, Xueguang Ma, Jimmy Lin, and Jamie Callan.
\newblock Precise zero-shot dense retrieval without relevance labels.
\newblock In Anna Rogers, Jordan Boyd-Graber, and Naoaki Okazaki, editors, \emph{Proceedings of the 61st Annual Meeting of the Association for Computational Linguistics (Volume 1: Long Papers)}, pages 1762--1777, Toronto, Canada, July 2023. Association for Computational Linguistics.
\newblock \doi{10.18653/v1/2023.acl-long.99}.
\newblock URL \url{https://aclanthology.org/2023.acl-long.99/}.

\bibitem[Gao et~al.(2021)Gao, Yao, and Chen]{gao-etal-2021-simcse}
Tianyu Gao, Xingcheng Yao, and Danqi Chen.
\newblock {S}im{CSE}: Simple contrastive learning of sentence embeddings.
\newblock In Marie-Francine Moens, Xuanjing Huang, Lucia Specia, and Scott Wen-tau Yih, editors, \emph{Proceedings of the 2021 Conference on Empirical Methods in Natural Language Processing}, pages 6894--6910, Online and Punta Cana, Dominican Republic, November 2021. Association for Computational Linguistics.
\newblock \doi{10.18653/v1/2021.emnlp-main.552}.
\newblock URL \url{https://aclanthology.org/2021.emnlp-main.552/}.

\bibitem[Gao et~al.(2024)Gao, Xiong, Gao, Jia, Pan, Bi, Dai, Sun, Wang, and Wang]{gao2024ragsurvey}
Yunfan Gao, Yun Xiong, Xinyu Gao, Kangxiang Jia, Jinliu Pan, Yuxi Bi, Yi~Dai, Jiawei Sun, Meng Wang, and Haofen Wang.
\newblock Retrieval-augmented generation for large language models: A survey.
\newblock \emph{arXiv:2312.10997}, 2024.
\newblock URL \url{https://arxiv.org/abs/2312.10997}.

\bibitem[Gusenbauer and Haddaway(2020)]{gusenbauer2020academic}
Michael Gusenbauer and Neal~R Haddaway.
\newblock Which academic search systems are suitable for systematic reviews or meta-analyses? evaluating retrieval qualities of google scholar, pubmed, and 26 other resources.
\newblock \emph{Research synthesis methods}, 11\penalty0 (2):\penalty0 181--217, 2020.

\bibitem[He et~al.(2025)He, Huang, Feng, Lin, Zhang, Li, and E]{he-etal-2025-pasa}
Yichen He, Guanhua Huang, Peiyuan Feng, Yuan Lin, Yuchen Zhang, Hang Li, and Weinan E.
\newblock {P}a{S}a: An {LLM} agent for comprehensive academic paper search.
\newblock In Wanxiang Che, Joyce Nabende, Ekaterina Shutova, and Mohammad~Taher Pilehvar, editors, \emph{Proceedings of the 63rd Annual Meeting of the Association for Computational Linguistics (Volume 1: Long Papers)}, pages 11663--11679, Vienna, Austria, July 2025. Association for Computational Linguistics.
\newblock ISBN 979-8-89176-251-0.
\newblock \doi{10.18653/v1/2025.acl-long.572}.
\newblock URL \url{https://aclanthology.org/2025.acl-long.572/}.

\bibitem[Karpukhin et~al.(2020)Karpukhin, Oğuz, Min, Lewis, Wu, Edunov, Chen, and tau Yih]{karpukhin2020densepassageretrievalopendomain}
Vladimir Karpukhin, Barlas Oğuz, Sewon Min, Patrick Lewis, Ledell Wu, Sergey Edunov, Danqi Chen, and Wen tau Yih.
\newblock Dense passage retrieval for open-domain question answering, 2020.
\newblock URL \url{https://arxiv.org/abs/2004.04906}.

\bibitem[Khattab and Zaharia(2020)]{colbert}
Omar Khattab and Matei Zaharia.
\newblock Col{BERT}: Efficient and effective passage search via contextualized late interaction over bert.
\newblock In \emph{Proceedings of the 43rd International ACM SIGIR Conference on Research and Development in Information Retrieval}, SIGIR '20, page 39–48, New York, NY, USA, 2020. Association for Computing Machinery.
\newblock ISBN 9781450380164.
\newblock \doi{10.1145/3397271.3401075}.
\newblock URL \url{https://doi.org/10.1145/3397271.3401075}.

\bibitem[Kingma and Ba(2015)]{kingma2015adam}
Diederik~P Kingma and Jimmy Ba.
\newblock Adam: A method for stochastic optimization.
\newblock \emph{International Conference on Learning Representations}, 2015.
\newblock URL \url{https://arxiv.org/abs/1412.6980}.

\bibitem[Larson et~al.(2019)Larson, Mahendran, Peper, Clarke, Lee, Hill, Kummerfeld, Leach, Laurenzano, Tang, and Mars]{larson2019clinc150}
Stefan Larson, Anish Mahendran, Joseph~J. Peper, Christopher Clarke, Andrew Lee, Parker Hill, Jonathan~K. Kummerfeld, Kevin Leach, Michael~A. Laurenzano, Lingjia Tang, and Jason Mars.
\newblock An evaluation dataset for intent classification and out-of-scope prediction, 2019.
\newblock URL \url{https://arxiv.org/abs/1909.02027}.

\bibitem[Lee et~al.(2025{\natexlab{a}})Lee, Roy, Xu, Raiman, Shoeybi, Catanzaro, and Ping]{lee2025nvembed}
Chankyu Lee, Rajarshi Roy, Mengyao Xu, Jonathan Raiman, Mohammad Shoeybi, Bryan Catanzaro, and Wei Ping.
\newblock {NV}-embed: Improved techniques for training {LLM}s as generalist embedding models.
\newblock In \emph{The Thirteenth International Conference on Learning Representations}, 2025{\natexlab{a}}.
\newblock URL \url{https://openreview.net/forum?id=lgsyLSsDRe}.

\bibitem[Lee et~al.(2025{\natexlab{b}})Lee, Chen, Dua, Cer, Shanbhogue, Naim, Ábrego, Li, Chen, Vera, Ren, Zhang, Salz, Boratko, Han, Chen, Huang, Rao, Suganthan, Han, Doumanoglou, Gupta, Moiseev, Yip, Jain, Baumgartner, Shahi, Gomez, Mariserla, Choi, Shah, Goenka, Chen, Xia, Chen, Duddu, Chen, Walker, Zhou, Ghiya, Gleicher, Gill, Dong, Seyedhosseini, Sung, Hoffmann, and Duerig]{lee2025geminiembeddinggeneralizableembeddings}
Jinhyuk Lee, Feiyang Chen, Sahil Dua, Daniel Cer, Madhuri Shanbhogue, Iftekhar Naim, Gustavo~Hernández Ábrego, Zhe Li, Kaifeng Chen, Henrique~Schechter Vera, Xiaoqi Ren, Shanfeng Zhang, Daniel Salz, Michael Boratko, Jay Han, Blair Chen, Shuo Huang, Vikram Rao, Paul Suganthan, Feng Han, Andreas Doumanoglou, Nithi Gupta, Fedor Moiseev, Cathy Yip, Aashi Jain, Simon Baumgartner, Shahrokh Shahi, Frank~Palma Gomez, Sandeep Mariserla, Min Choi, Parashar Shah, Sonam Goenka, Ke~Chen, Ye~Xia, Koert Chen, Sai Meher~Karthik Duddu, Yichang Chen, Trevor Walker, Wenlei Zhou, Rakesh Ghiya, Zach Gleicher, Karan Gill, Zhe Dong, Mojtaba Seyedhosseini, Yunhsuan Sung, Raphael Hoffmann, and Tom Duerig.
\newblock Gemini embedding: Generalizable embeddings from gemini.
\newblock \emph{arXiv:2503.07891}, 2025{\natexlab{b}}.
\newblock URL \url{https://arxiv.org/abs/2503.07891}.

\bibitem[Lewis et~al.(2020)Lewis, Perez, Piktus, Petroni, Karpukhin, Goyal, K\"{u}ttler, Lewis, Yih, Rockt\"{a}schel, Riedel, and Kiela]{lewis2020rag}
Patrick Lewis, Ethan Perez, Aleksandra Piktus, Fabio Petroni, Vladimir Karpukhin, Naman Goyal, Heinrich K\"{u}ttler, Mike Lewis, Wen-tau Yih, Tim Rockt\"{a}schel, Sebastian Riedel, and Douwe Kiela.
\newblock Retrieval-augmented generation for knowledge-intensive nlp tasks.
\newblock In \emph{Proceedings of the 34th International Conference on Neural Information Processing Systems}, NIPS '20, Red Hook, NY, USA, 2020. Curran Associates Inc.
\newblock ISBN 9781713829546.

\bibitem[Li et~al.(2025)Li, Qin, Xiao, Chen, Luo, Lian, Shao, and Liu]{li2025making}
Chaofan Li, Minghao Qin, Shitao Xiao, Jianlyu Chen, Kun Luo, Defu Lian, Yingxia Shao, and Zheng Liu.
\newblock Making text embedders few-shot learners.
\newblock In \emph{The Thirteenth International Conference on Learning Representations}, 2025.
\newblock URL \url{https://openreview.net/forum?id=wfLuiDjQ0u}.

\bibitem[Liu et~al.(2019)Liu, Li, and Lin]{liu2019review}
Jiao Liu, Yanling Li, and Min Lin.
\newblock Review of intent detection methods in the human-machine dialogue system.
\newblock In \emph{Journal of Physics: Conference Series}, volume 1267. IOP Publishing, 2019.

\bibitem[Muennighoff et~al.(2023)Muennighoff, Tazi, Magne, and Reimers]{muennighoff-etal-2023-mteb}
Niklas Muennighoff, Nouamane Tazi, Loic Magne, and Nils Reimers.
\newblock {MTEB}: Massive text embedding benchmark.
\newblock In Andreas Vlachos and Isabelle Augenstein, editors, \emph{Proceedings of the 17th Conference of the European Chapter of the Association for Computational Linguistics}, pages 2014--2037, Dubrovnik, Croatia, May 2023. Association for Computational Linguistics.
\newblock \doi{10.18653/v1/2023.eacl-main.148}.
\newblock URL \url{https://aclanthology.org/2023.eacl-main.148/}.

\bibitem[Nooten et~al.(2025)Nooten, Kosar, Pauw, and Daelemans]{vannooten2025sizedoesfitall}
Jens~Van Nooten, Andriy Kosar, Guy~De Pauw, and Walter Daelemans.
\newblock One size does not fit all: Exploring variable thresholds for distance-based multi-label text classification, 2025.
\newblock URL \url{https://arxiv.org/abs/2510.11160}.

\bibitem[OpenAI et~al.(2025)OpenAI, :, Agarwal, Ahmad, Ai, Altman, Applebaum, Arbus, Arora, Bai, Baker, Bao, Barak, Bennett, Bertao, Brett, Brevdo, Brockman, Bubeck, Chang, Chen, Chen, Cheung, Clark, Cook, Dukhan, Dvorak, Fives, Fomenko, Garipov, Georgiev, Glaese, Gogineni, Goucher, Gross, Guzman, Hallman, Hehir, Heidecke, Helyar, Hu, Huet, Huh, Jain, Johnson, Koch, Kofman, Kundel, Kwon, Kyrylov, Le, Leclerc, Lennon, Lessans, Lezcano-Casado, Li, Li, Lin, Liss, Lily, Liu, Liu, Lu, Lu, Martinovic, McCallum, McGrath, McKinney, McLaughlin, Mei, Mostovoy, Mu, Myles, Neitz, Nichol, Pachocki, Paino, Palmie, Pantuliano, Parascandolo, Park, Pathak, Paz, Peran, Pimenov, Pokrass, Proehl, Qiu, Raila, Raso, Ren, Richardson, Robinson, Rotsted, Salman, Sanjeev, Schwarzer, Sculley, Sikchi, Simon, Singhal, Song, Stuckey, Sun, Tillet, Toizer, Tsimpourlas, Vyas, Wallace, Wang, Wang, Watkins, Weil, Wendling, Whinnery, Whitney, Wong, Yang, Yang, Yasunaga, Ying, Zaremba, Zhan, Zhang, Zhang, Zhang, and
  Zhao]{openai2025gptoss120bgptoss20bmodel}
OpenAI, :, Sandhini Agarwal, Lama Ahmad, Jason Ai, Sam Altman, Andy Applebaum, Edwin Arbus, Rahul~K. Arora, Yu~Bai, Bowen Baker, Haiming Bao, Boaz Barak, Ally Bennett, Tyler Bertao, Nivedita Brett, Eugene Brevdo, Greg Brockman, Sebastien Bubeck, Che Chang, Kai Chen, Mark Chen, Enoch Cheung, Aidan Clark, Dan Cook, Marat Dukhan, Casey Dvorak, Kevin Fives, Vlad Fomenko, Timur Garipov, Kristian Georgiev, Mia Glaese, Tarun Gogineni, Adam Goucher, Lukas Gross, Katia~Gil Guzman, John Hallman, Jackie Hehir, Johannes Heidecke, Alec Helyar, Haitang Hu, Romain Huet, Jacob Huh, Saachi Jain, Zach Johnson, Chris Koch, Irina Kofman, Dominik Kundel, Jason Kwon, Volodymyr Kyrylov, Elaine~Ya Le, Guillaume Leclerc, James~Park Lennon, Scott Lessans, Mario Lezcano-Casado, Yuanzhi Li, Zhuohan Li, Ji~Lin, Jordan Liss, Lily, Liu, Jiancheng Liu, Kevin Lu, Chris Lu, Zoran Martinovic, Lindsay McCallum, Josh McGrath, Scott McKinney, Aidan McLaughlin, Song Mei, Steve Mostovoy, Tong Mu, Gideon Myles, Alexander Neitz, Alex Nichol, Jakub
  Pachocki, Alex Paino, Dana Palmie, Ashley Pantuliano, Giambattista Parascandolo, Jongsoo Park, Leher Pathak, Carolina Paz, Ludovic Peran, Dmitry Pimenov, Michelle Pokrass, Elizabeth Proehl, Huida Qiu, Gaby Raila, Filippo Raso, Hongyu Ren, Kimmy Richardson, David Robinson, Bob Rotsted, Hadi Salman, Suvansh Sanjeev, Max Schwarzer, D.~Sculley, Harshit Sikchi, Kendal Simon, Karan Singhal, Yang Song, Dane Stuckey, Zhiqing Sun, Philippe Tillet, Sam Toizer, Foivos Tsimpourlas, Nikhil Vyas, Eric Wallace, Xin Wang, Miles Wang, Olivia Watkins, Kevin Weil, Amy Wendling, Kevin Whinnery, Cedric Whitney, Hannah Wong, Lin Yang, Yu~Yang, Michihiro Yasunaga, Kristen Ying, Wojciech Zaremba, Wenting Zhan, Cyril Zhang, Brian Zhang, Eddie Zhang, and Shengjia Zhao.
\newblock gpt-oss-120b and gpt-oss-20b model card, 2025.
\newblock URL \url{https://arxiv.org/abs/2508.10925}.

\bibitem[Reimers and Gurevych(2019)]{reimers-gurevych-2019-sentence}
Nils Reimers and Iryna Gurevych.
\newblock Sentence-{BERT}: Sentence embeddings using {S}iamese {BERT}-networks.
\newblock In Kentaro Inui, Jing Jiang, Vincent Ng, and Xiaojun Wan, editors, \emph{Proceedings of the 2019 Conference on Empirical Methods in Natural Language Processing and the 9th International Joint Conference on Natural Language Processing (EMNLP-IJCNLP)}, pages 3982--3992, Hong Kong, China, November 2019. Association for Computational Linguistics.
\newblock \doi{10.18653/v1/D19-1410}.
\newblock URL \url{https://aclanthology.org/D19-1410/}.

\bibitem[Su et~al.(2023)Su, Shi, Kasai, Wang, Hu, Ostendorf, Yih, Smith, Zettlemoyer, and Yu]{su-etal-2023-one}
Hongjin Su, Weijia Shi, Jungo Kasai, Yizhong Wang, Yushi Hu, Mari Ostendorf, Wen-tau Yih, Noah~A. Smith, Luke Zettlemoyer, and Tao Yu.
\newblock One embedder, any task: Instruction-finetuned text embeddings.
\newblock In Anna Rogers, Jordan Boyd-Graber, and Naoaki Okazaki, editors, \emph{Findings of the Association for Computational Linguistics: ACL 2023}, pages 1102--1121, Toronto, Canada, July 2023. Association for Computational Linguistics.
\newblock \doi{10.18653/v1/2023.findings-acl.71}.
\newblock URL \url{https://aclanthology.org/2023.findings-acl.71/}.

\bibitem[Su et~al.(2025)Su, Yen, Xia, Shi, Muennighoff, yu~Wang, Haisu, Shi, Siegel, Tang, Sun, Yoon, Arik, Chen, and Yu]{su2025bright}
Hongjin Su, Howard Yen, Mengzhou Xia, Weijia Shi, Niklas Muennighoff, Han yu~Wang, Liu Haisu, Quan Shi, Zachary~S Siegel, Michael Tang, Ruoxi Sun, Jinsung Yoon, Sercan~O Arik, Danqi Chen, and Tao Yu.
\newblock {BRIGHT}: A realistic and challenging benchmark for reasoning-intensive retrieval.
\newblock In \emph{The Thirteenth International Conference on Learning Representations}, 2025.
\newblock URL \url{https://openreview.net/forum?id=ykuc5q381b}.

\bibitem[Sun et~al.(2024)Sun, Shi, Long, Yan, Ma, Liu, Cao, Yin, and Ren]{sun-etal-2024-mair}
Weiwei Sun, Zhengliang Shi, Wu~Jiu Long, Lingyong Yan, Xinyu Ma, Yiding Liu, Min Cao, Dawei Yin, and Zhaochun Ren.
\newblock {MAIR}: A massive benchmark for evaluating instructed retrieval.
\newblock In Yaser Al-Onaizan, Mohit Bansal, and Yun-Nung Chen, editors, \emph{Proceedings of the 2024 Conference on Empirical Methods in Natural Language Processing}, pages 14044--14067, Miami, Florida, USA, November 2024. Association for Computational Linguistics.
\newblock \doi{10.18653/v1/2024.emnlp-main.778}.
\newblock URL \url{https://aclanthology.org/2024.emnlp-main.778/}.

\bibitem[Sung et~al.(2023)Sung, Park, Kang, Chen, and Lee]{sung-etal-2023-optimizing}
Mujeen Sung, Jungsoo Park, Jaewoo Kang, Danqi Chen, and Jinhyuk Lee.
\newblock Optimizing test-time query representations for dense retrieval.
\newblock In Anna Rogers, Jordan Boyd-Graber, and Naoaki Okazaki, editors, \emph{Findings of the Association for Computational Linguistics: ACL 2023}, pages 5731--5746, Toronto, Canada, July 2023. Association for Computational Linguistics.
\newblock URL \url{https://aclanthology.org/2023.findings-acl.354/}.

\bibitem[Thakur et~al.(2021)Thakur, Reimers, R{\"u}ckl{\'e}, Srivastava, and Gurevych]{thakur2021beir}
Nandan Thakur, Nils Reimers, Andreas R{\"u}ckl{\'e}, Abhishek Srivastava, and Iryna Gurevych.
\newblock {BEIR}: A heterogeneous benchmark for zero-shot evaluation of information retrieval models.
\newblock In \emph{Thirty-fifth Conference on Neural Information Processing Systems Datasets and Benchmarks Track (Round 2)}, 2021.
\newblock URL \url{https://openreview.net/forum?id=wCu6T5xFjeJ}.

\bibitem[Uzan et~al.(2025)Uzan, Yehudai, Pony, Shnarch, and Gera]{uzan2025guidedqueryrefinementmultimodal}
Omri Uzan, Asaf Yehudai, Roi Pony, Eyal Shnarch, and Ariel Gera.
\newblock Guided query refinement: Multimodal hybrid retrieval with test-time optimization.
\newblock \emph{arXiv:2510.05038}, 2025.
\newblock URL \url{https://arxiv.org/abs/2510.05038}.

\bibitem[Wang et~al.(2024)Wang, Yang, Huang, Yang, Majumder, and Wei]{wang-etal-2024-improving-text}
Liang Wang, Nan Yang, Xiaolong Huang, Linjun Yang, Rangan Majumder, and Furu Wei.
\newblock Improving text embeddings with large language models.
\newblock In Lun-Wei Ku, Andre Martins, and Vivek Srikumar, editors, \emph{Proceedings of the 62nd Annual Meeting of the Association for Computational Linguistics (Volume 1: Long Papers)}, pages 11897--11916, Bangkok, Thailand, August 2024. Association for Computational Linguistics.
\newblock \doi{10.18653/v1/2024.acl-long.642}.
\newblock URL \url{https://aclanthology.org/2024.acl-long.642/}.

\bibitem[Weller et~al.(2025)Weller, Chang, MacAvaney, Lo, Cohan, Van~Durme, Lawrie, and Soldaini]{weller-etal-2025-followir}
Orion Weller, Benjamin Chang, Sean MacAvaney, Kyle Lo, Arman Cohan, Benjamin Van~Durme, Dawn Lawrie, and Luca Soldaini.
\newblock {F}ollow{IR}: Evaluating and teaching information retrieval models to follow instructions.
\newblock In Luis Chiruzzo, Alan Ritter, and Lu~Wang, editors, \emph{Proceedings of the 2025 Conference of the Nations of the Americas Chapter of the Association for Computational Linguistics: Human Language Technologies (Volume 1: Long Papers)}, pages 11926--11942, Albuquerque, New Mexico, April 2025. Association for Computational Linguistics.
\newblock ISBN 979-8-89176-189-6.
\newblock \doi{10.18653/v1/2025.naacl-long.597}.
\newblock URL \url{https://aclanthology.org/2025.naacl-long.597/}.

\bibitem[Yu et~al.(2021)Yu, Xiong, and Callan]{yu2021improving}
HongChien Yu, Chenyan Xiong, and Jamie Callan.
\newblock Improving query representations for dense retrieval with pseudo relevance feedback.
\newblock In \emph{Proceedings of the 30th ACM International Conference on Information \& Knowledge Management}, CIKM '21, page 3592–3596, New York, NY, USA, 2021. Association for Computing Machinery.
\newblock ISBN 9781450384469.
\newblock \doi{10.1145/3459637.3482124}.
\newblock URL \url{https://doi.org/10.1145/3459637.3482124}.

\bibitem[Zhang et~al.(2025)Zhang, Li, Long, Zhang, Lin, Yang, Xie, Yang, Liu, Lin, Huang, and Zhou]{qwen3embedding}
Yanzhao Zhang, Mingxin Li, Dingkun Long, Xin Zhang, Huan Lin, Baosong Yang, Pengjun Xie, An~Yang, Dayiheng Liu, Junyang Lin, Fei Huang, and Jingren Zhou.
\newblock Qwen3 embedding: Advancing text embedding and reranking through foundation models.
\newblock \emph{arXiv:2506.05176}, 2025.
\newblock URL \url{https://arxiv.org/abs/2506.05176}.

\bibitem[Zhang et~al.(2022)Zhang, Strubell, and Hovy]{zhang-etal-2022-survey}
Zhisong Zhang, Emma Strubell, and Eduard Hovy.
\newblock A survey of active learning for natural language processing.
\newblock In Yoav Goldberg, Zornitsa Kozareva, and Yue Zhang, editors, \emph{Proceedings of the 2022 Conference on Empirical Methods in Natural Language Processing}, pages 6166--6190, Abu Dhabi, United Arab Emirates, December 2022. Association for Computational Linguistics.
\newblock \doi{10.18653/v1/2022.emnlp-main.414}.
\newblock URL \url{https://aclanthology.org/2022.emnlp-main.414}.

\bibitem[Zhuang et~al.(2024)Zhuang, Zhuang, Koopman, and Zuccon]{Zhuang_2024}
Shengyao Zhuang, Honglei Zhuang, Bevan Koopman, and Guido Zuccon.
\newblock A setwise approach for effective and highly efficient zero-shot ranking with large language models.
\newblock In \emph{Proceedings of the 47th International ACM SIGIR Conference on Research and Development in Information Retrieval}, SIGIR 2024, page 38–47. ACM, July 2024.
\newblock \doi{10.1145/3626772.3657813}.
\newblock URL \url{http://dx.doi.org/10.1145/3626772.3657813}.

\bibitem[Zuccon et~al.(2025)Zuccon, Zhuang, and Ma]{zuccon2025r2llms}
Guido Zuccon, Shengyao Zhuang, and Xueguang Ma.
\newblock R2{LLM}s: Retrieval and ranking with {LLM}s.
\newblock In \emph{Proceedings of the 48th International ACM SIGIR Conference on Research and Development in Information Retrieval}, pages 4106--4109, 2025.
\newblock URL \url{https://ielab.io/tutorials/r2llms.html}.

\end{thebibliography}


\appendix

\newpage
\appendix
\onecolumn

\section{Dataset Details} \label{app:datasets}
The details of the datasets used in this work are provided in Table~\ref{tab:datasets}.

\begin{table}[h]
    \centering
    \caption{Details of datasets used in our experiments.}
    \label{tab:datasets}
    \resizebox{\textwidth}{!}{%

    \begin{tabular}{l|p{8.5cm}|c|c}
        \textbf{Dataset} & \textbf{Description} & \textbf{\# Queries} & \textbf{\# Positives/query}\\
        \toprule
        \realscholarlong{} \citep{he-etal-2025-pasa} &
        Real-world search queries over a large corpus of computer science papers from arXiv. Expert annotators provided comprehensive relevance labels. &
        $50$ & $14.6$ \\
        \midrule
        \clinc{} \citep{larson2019clinc150} &
        Intent‑classification benchmark with $150$ intents across $10$ domains, featuring customer service queries collected via crowdsourcing. &
        $150$ & $100$ \\
        \midrule
        \kpm{} \citep{friedman-etal-2021-overview} &
        Key-point–matching dataset from the $2021$ KPA shared task. Each query corresponds to a key point, and the objective is accurate pairwise matching between key points and arguments. &
        $243$ & $20.6$\\
        \midrule
        \followir{} \citep{weller-etal-2025-followir} &
        An information-retrieval benchmark derived from detailed TREC relevance narratives. Includes pairs of very similar queries with nuanced differences in the instruction. &
        $208$ & $30$ \\
        \midrule
        \banking{} \citep{casanueva-etal-2020-efficient} &
        A task-oriented banking-domain dataset that classifies user requests into $77$ fine-grained intent categories. &
        $77$ & $169.7$ \\
        \midrule
        \textsc{NFCorpus} \citep{boteva2016full} &
        A retrieval dataset with over $3{,}000$ lay queries, used to identify relevant medical literature articles. &
        $323$ & $38.2$ \\

        \bottomrule
    \end{tabular}
    }
\end{table}

\paragraph{\clinc{} queries} Given that the original \clinc{} intent labels are difficult to interpret, as a preliminary step we use GPT-OSS-120B \citep{openai2025gptoss120bgptoss20bmodel} to rephrase the intents as queries in natural language. The LLM is given a sample of 10 positive instances for phrasing the new query, using the following 1-shot prompt:\\ 
\texttt{Suggest an intent description for the following user queries:\\
how do i find my credit score\\
help me locate my credit score \\
...\\
if i want my credit score, how do i find it \\
Intent description: Requests to view or obtain a user's credit score, including asking how to find it.}

\section{Comparison to Prior Work} \label{app:prior_work}
Table~\ref{tab:prior_work} compares our work with prior query refinement efforts, ReFit~\citep{gangireddy2025rerankerrelevancefeedback} and GQR~\citep{uzan2025guidedqueryrefinementmultimodal}.

The central difference from prior works is in the overall task setup -- we extend this from general-purpose retrieval to varied LLM-guided binary separation tasks, covering classic classification scenarios as well as challenging and nuanced relevance decisions. This makes the paradigm relevant to a broader community of use cases and practitioners, extending the notion of embedding adaptability by harnessing the zero-shot capabilities of LLMs.

\begin{table*}[h]
\centering
\caption{Comparison to prior query refinement works.} \label{tab:prior_work}

\resizebox{\textwidth}{!}{%

\begin{tabular}{llllll}
\toprule
Method	& Target Tasks	& Feedback Model & 	Primary Model &	Loss Function	& Score Normalization \\
\midrule
\textsc{ReFIT} &	General-purpose	& Cross-Encoder	& Text Encoder	& \textit{KL-divergence with} & Min–Max \\
& Retrieval & Reranker & & \textit{feedback model}	& + Softmax (T = 2) \\
\midrule
\textsc{GQR}	& Visual Document	& Text Encoders	& Visual Encoders & 	KL-divergence with & \textit{Softmax} \\
& Retrieval & & & consensus (average) distribution	& \\
\midrule
\textsc{Ours} &	\textbf{Nuanced Classification} & \textbf{LLM}	& Text Encoders	& \textit{KL-divergence with} & \textit{Softmax} \\
& \textbf{and Search} & & & \textit{feedback model} & \\
\bottomrule
\end{tabular}

}

\end{table*}

\subsection{Empirical Comparison to ReFit}
As stated above, our key contribution lies in the task setup and not in the precise refinement algorithm. Nevertheless, for completeness, we also conduct an empirical comparison with the refinement setup from ReFit~\citep{gangireddy2025rerankerrelevancefeedback}.\footnote{We do not compare to GQR given the substantial differences in setup -- they look at a multimodal scenario of combining a text and vision encoder.} Namely, we run experiments on the datasets with the ReFIT setup, using their choice of cross-encoder, score normalization approach and hyperparameters. Since their original choice of learning rate led to poor results, we also ran tests with the same learning rate as our setup.

Table~\ref{tab:refit} depicts the results, showing that the specific implementation choices made in ReFit (including a cross-encoder tuned on MS-Marco, normalization with temperature, and the learning rate) do not transfer well to our scenario.

\begin{table*}
\centering
\caption{Mean average precision (MAP) results across queries, comparing our query refinement setup with that from ReFit~\cite{gangireddy2025rerankerrelevancefeedback}.} \label{tab:refit}

\resizebox{\textwidth}{!}{%

\begin{tabular}{lllllll}
\toprule
Model & Variant & \realscholar{} & \clinc{} & \kpm{}  & \followir{} & Average \\
\midrule
\qwensmall{} & Original & .54 & .59 & .67 & .41 & .55 \\
& ReFIT & .47 & .52  & .65 & .39 & .51 \\
& ReFIT (LR=1e‑4) & .54 & .61  & .67 & .41 & .56 \\
& Ours  & .65 & .74  & .79 & .46 & \textbf{.66} \\
\midrule
\qwenlarge{} & Original & .61 & .86 & .77 & .45 & .67 \\
& ReFIT & .47 & .59  & .66 & .37 & .52 \\
& ReFIT (LR=1e‑4) & .59 & .86  & .72 & .43 & .65 \\
& Ours  & .70 & .88  & .83 & .49 & \textbf{.72} \\
\midrule
\nemotron{} & Original & .54 & .80 & .59 & .45 & .59 \\
& ReFIT & .40 & .46 & .59 & .36 & .45 \\
& ReFIT (LR=1e‑4) & .55 & .80 & .60 & .42 & .59 \\
& Ours  & .67 & .85  & .74 & .47 & \textbf{.68} \\
\midrule
\linq{} & Original & .51 & .76 & .67 & .43 & .59 \\
& ReFIT & .53 & .76  & .67 & .43 & .60 \\
& ReFIT (LR=1e‑4) & .51 & .76  & .67 & .43 & .59 \\
& Ours  & .51 & .76 & .67 & .43 & \textbf{.59} \\
\midrule
\efivemistral{} & Original & .48 & .69 & .63 & .41 & .55 \\
& ReFIT & .36 & .35 & .60 & .35 & .42 \\
& ReFIT (LR=1e‑4) & .48 & .71 & .63 & .41 & .56 \\
& Ours  & .60 & .78 & .77 & .46 & \textbf{.65} \\
\bottomrule
\end{tabular}
}

\end{table*}

\section{Additional Results} \label{app:additional}
Table~\ref{tab:results_app_full} depicts the full mean average precision results across all datasets.

Table~\ref{tab:pvalues} shows the statistical significance testing for the improvement in average precision.

Table~\ref{tab:llms} compares query refinement results using different LLM teachers.

Table~\ref{tab:perf_correlations} depicts correlations between feedback quality and query refinement gains.

\subsection{Refinement with HyDE} \label{app:hyde}
A popular approach for improving zero-shot retrieval performance of embedding models is HyDE (Hypothetical Document Embeddings, \citealp{gao-etal-2023-hyde}). In this approach, an LLM is given the text of a user query, and is asked to generate a ``hypothetical'' document that would be relevant for this query. This fake LLM-generated document is then used as the query for retrieval. The intuition is that this can help bridge the gap between the nature of the queries and that of the corpus documents, making it easier to capture query-document similarities with embedding representations.

Thus, we explore the relationship between the HyDE paradigm of adaptating query \textit{texts} and our approach for adapting query \textit{embeddings}. Specifically, we conduct experiments comparing corpus ranking quality with four query settings: calculating similarity using the original embedded query (\textit{original}), a hypothetical-document query that was generated by an LLM (\textit{HyDE}), the refined embedding of the original query (\textit{optimized}), and the refined embedding of the HyDE query (\textit{HyDE optimized}).

In Table~\ref{tab:hyde_results} we see that HyDE does not have a positive effect in these datasets.
While this detrimental effect may seem surprising, one possible explanation is that HyDE introduces noise by relying on a generated document that is not perfectly aligned with the corpus, leading to embedding drift. Moreover, the modern embedding models used here are already trained with instruction‑following and retrieval‑oriented objectives that address some of the issues that HyDE was designed to solve.

Applying our query embedding refinement approach consistently improves results compared to the initial query embeddings, both for the original query texts and for the HyDE-style queries.

\section{Instructions and Prompts}

\subsection{Embedding Instructions} \label{app:embed_instructions}
As recommended in documentation of the various embedding models, we embed the query text as part of a query template that also includes a short task instruction:
{\small\verb|Instruct: {instruction}\nQuery: {query}|}

Accordingly, we also phrase dedicated task instructions for each of the datasets. In the box below, the top instruction is the default one we used for all the main experiments. 

\begin{tcolorbox}[title=Dataset Instruction Variants, colback=gray!5, colframe=black]

\textbf{\clinc{}}\\
-- \textit{Given an intent description, identify utterances that express this intent.}\\
-- Retrieve user utterances that align with the specific intent description provided, distinguishing them from out-of-scope inputs.\\

\textbf{\kpm{}}\\
-- \textit{Given an argument, identify utterances from debates that convey this argument.}\\
-- Retrieve debate utterances that semantically match and explicitly express the core argument provided.\\

\textbf{\followir{}}\\
-- \textit{Retrieve text based on a fine-grained user query.}\\

\textbf{\realscholarlong{}}\\
-- \textit{Given a fine-grained scientific inquiry, retrieve academic publications that strictly align with the specified research topic, methodology, and constraints.}\\
-- Retrieve academic publications that comprehensively address the research topic, methodology, and specific constraints detailed in the scientific inquiry.\\

\textbf{\nfcorpus{}}\\
-- \textit{Given a question, retrieve relevant documents that best answer the question.}\\
-- Retrieve relevant medical documents and passages that directly answer the user's health-related question.\\

\textbf{\banking{}}\\
-- \textit{Given an intent category, identify utterances that express this intent.}\\
-- Retrieve banking customer service utterances that clearly express the given intent category.\\

\end{tcolorbox}

\paragraph{Sensitivity to instructions}

Given our interest in utilizing general-purpose embedding models for varied tasks, we also examine the effect of the choice of instructions and templates.
Thus, we perform experiments comparing the two task instruction choices above, in combination with several query \textit{template} options:
\begin{enumerate}[nosep,label={$\bullet$},wide=5pt]
\item \textbf{Default}: {\small\verb|Instruct: {instruction}\nQuery: {query}|}
\item \textbf{No instruction}: {\small\verb|{query}|}
\item \textbf{Instruction prefix}: {\small\verb|{instruction}\n{query}|}
\item \textbf{``Task'' template}: {\small\verb|Task: {instruction}\n\n{query}|}
\item \textbf{``Retrieve'' template}: {\small\verb|Retrieve relevant documents. {instruction}\nQuery: {query}|}
\end{enumerate}

\begin{figure*}[ht]
    \centering

\subfloat[\clinc{}]{\includegraphics[width=.49\textwidth]{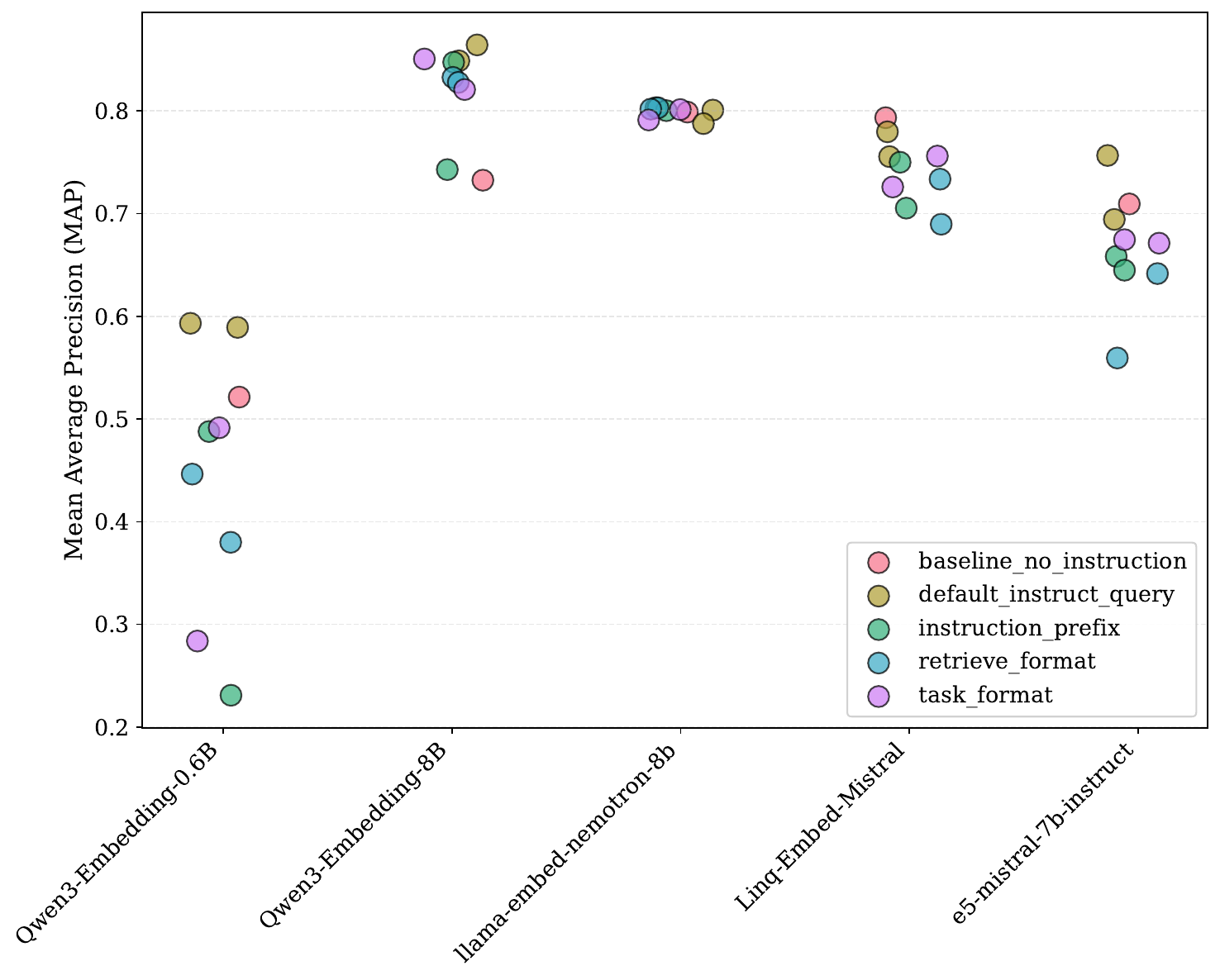} 
}
\subfloat[\kpm{}]{
\includegraphics[width=.49\textwidth]{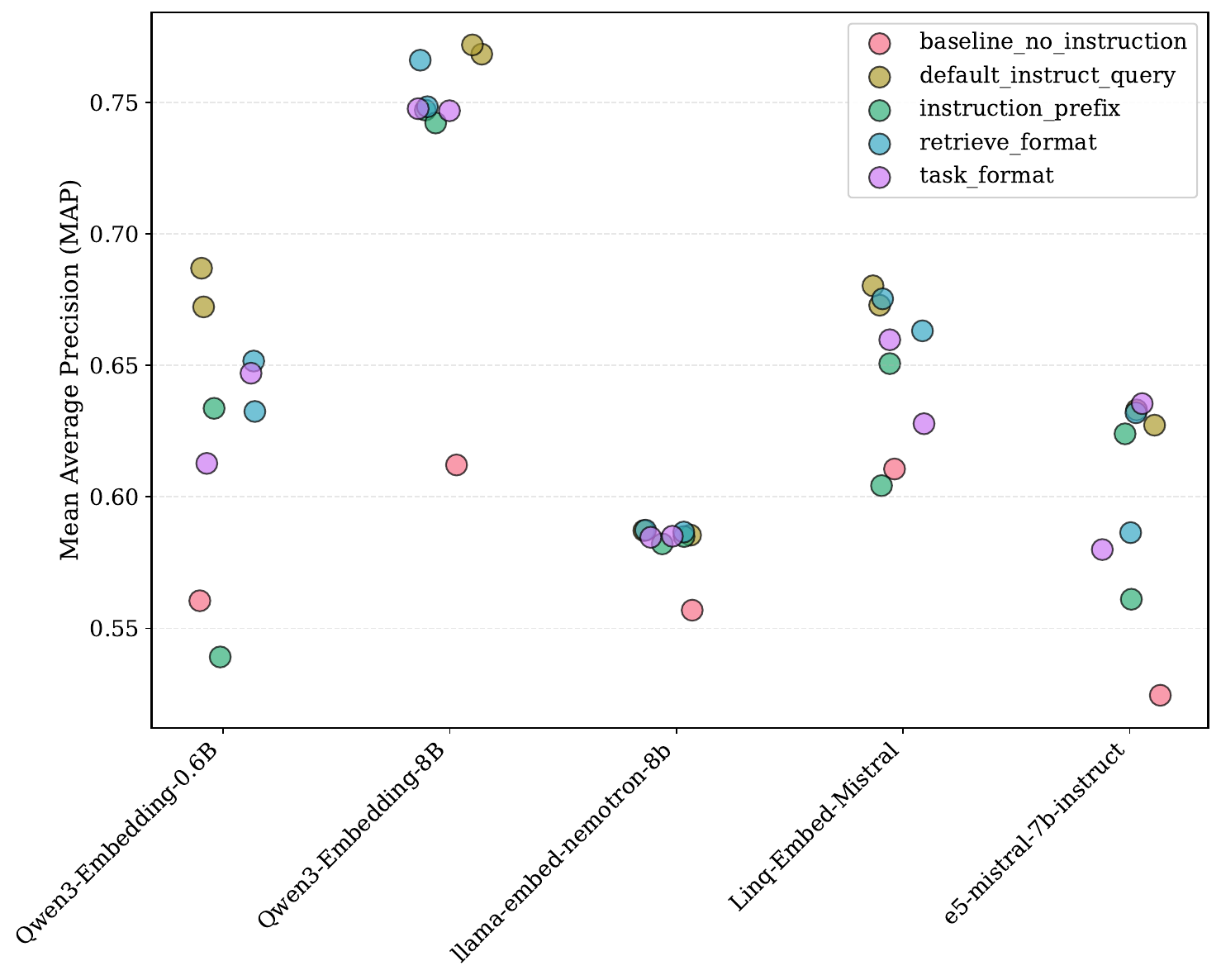} 
}

\vspace{0.5cm}
\subfloat[\realscholarlong{}]{
\includegraphics[width=.49\textwidth]{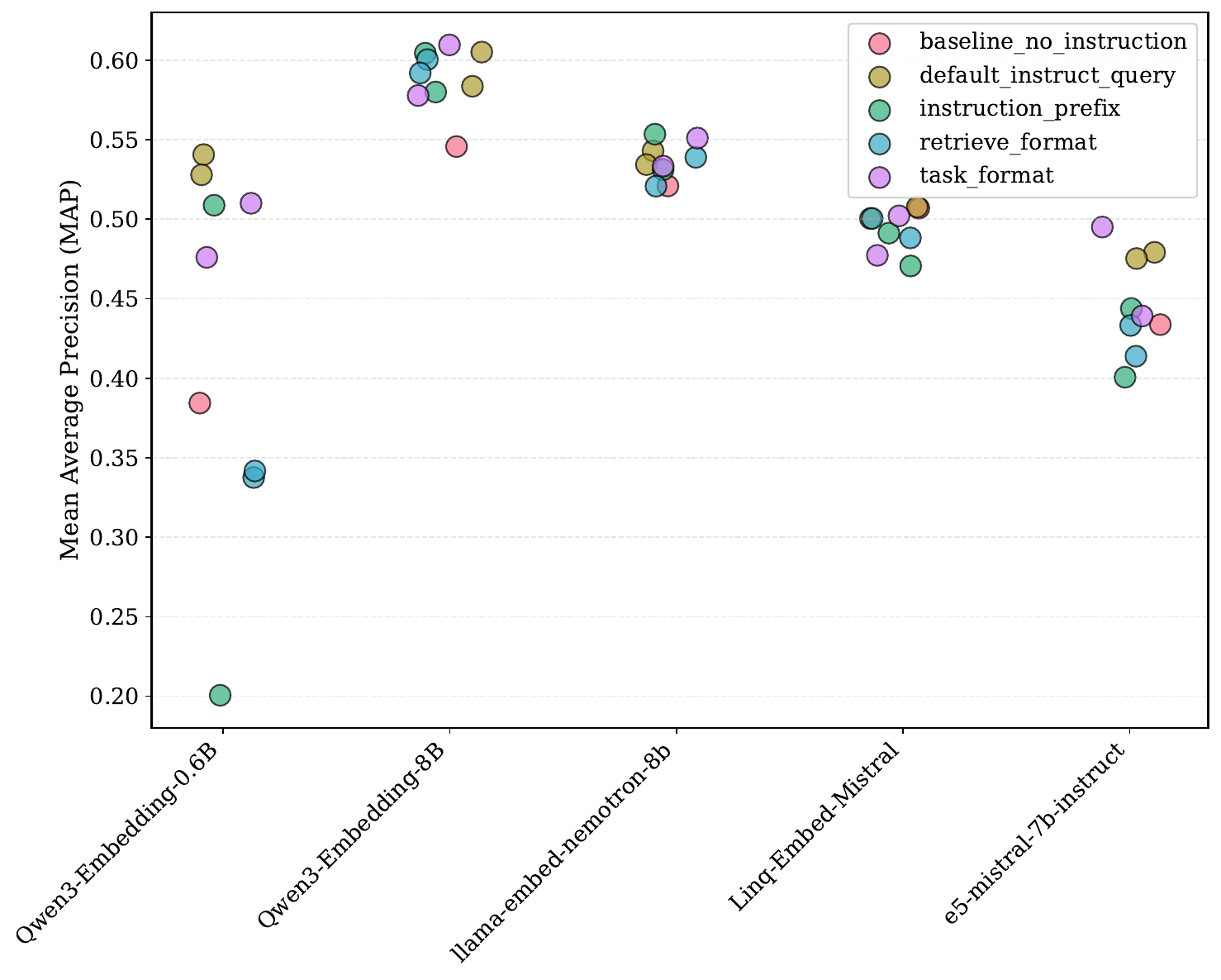} 
}
\subfloat[\banking{}]{
\includegraphics[width=.49\textwidth]{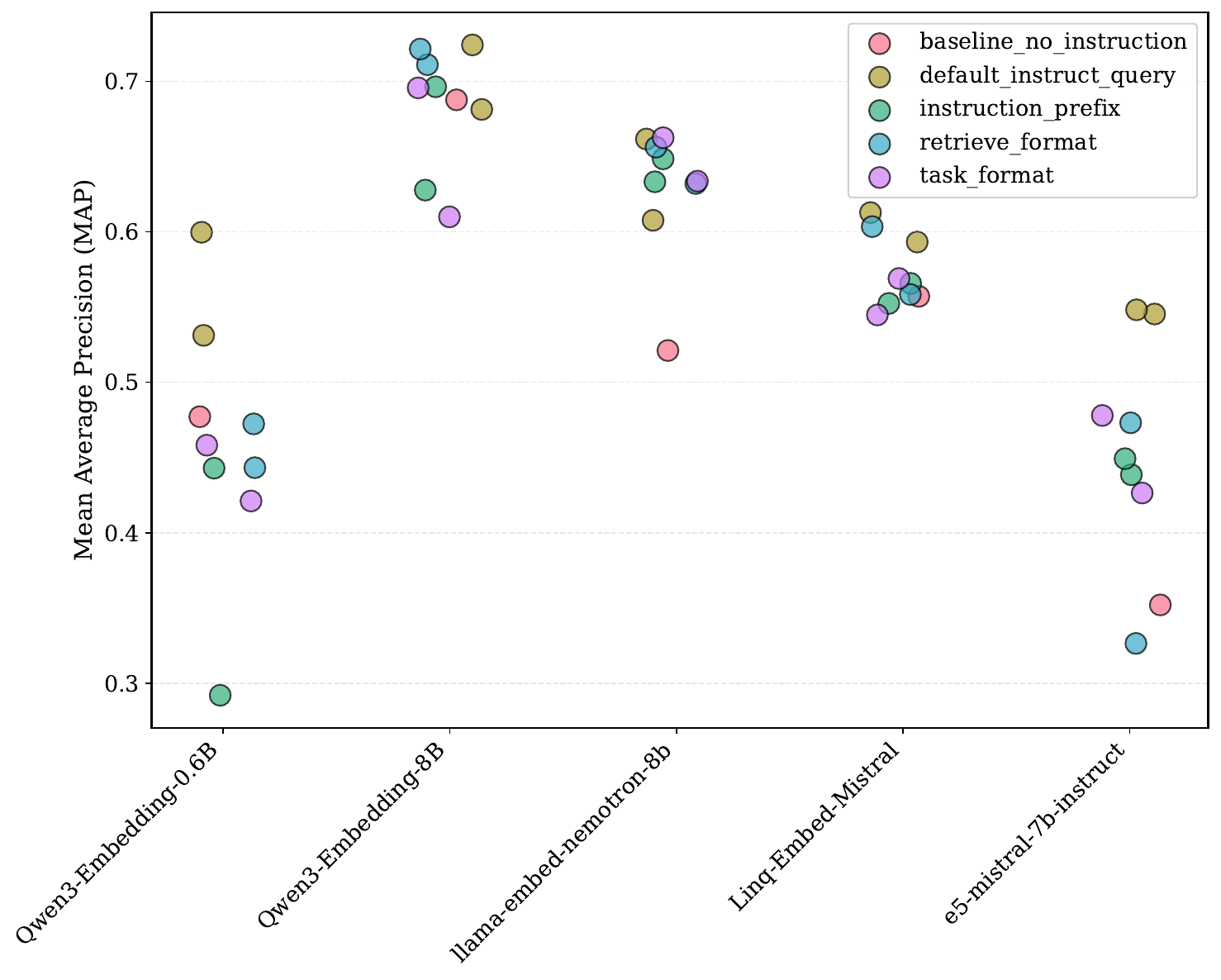} 
}

\caption{Comparing embedding model base performance when using different query templates and query task instructions.} \label{fig:instruct}
\end{figure*}

The results of this analysis are presented in Figure~\ref{fig:instruct}. We see that some models are more brittle than others. In particular, the smallest model in the group, \textit{\qwensmall{}}, is very sensitive to varying templates and instructions, much more than its 8B counterpart. \textit{\nemotron{}} is the most robust embedding model in this respect. We also see that the default query template, which is the one we use here and the one recommended by model documentation, indeed works well. It also appears that with these models, using no task instruction can be somewhat detrimental -- in particular to specialized tasks like \kpm{}.

\begin{table*}
\centering
\caption{Mean average precision (MAP) results across queries. Here we compare four settings for the query representations: embedding of the original query, embedding of a HyDE query, and optimized versions of these queries after applying LLM-guided query refinement (\S\ref{ssec:optimization}).} \label{tab:hyde_results}

\resizebox{\textwidth}{!}{%
\begin{tabular}{lllllllll}
\toprule
 &  & \realscholar{} & \clinc{} & \kpm{} & \followir{} & \nfcorpus{} & \banking{} & Average \\
Embedding Model & Setting &  &  &  &  &  &  &  \\
\midrule
\multirow[t]{5}{*}{\qwensmall{}} & Original & .54 & .59 & .67 & .41 & .20 & .53 & .49 \\
 & Rerank-only & .62 {\scriptsize (+15.5\%)} & .60 {\scriptsize (+1.3\%)} & .73 {\scriptsize (+8.0\%)} & .45 {\scriptsize (+8.2\%)} & .21 {\scriptsize (+8.0\%)} & .53 {\scriptsize (+0.2\%)} & .52 {\scriptsize (+6.7\%)} \\
 & Optimized & .65 {\scriptsize (+20.7\%)} & .74 {\scriptsize (+25.5\%)} & .79 {\scriptsize (+17.2\%)} & .46 {\scriptsize (+11.3\%)} & .23 {\scriptsize (+16.6\%)} & .60 {\scriptsize (+12.4\%)} & .58 {\scriptsize (+17.8\%)} \\
 & HyDE & .55 {\scriptsize (+2.1\%)} & .68 {\scriptsize (+16.0\%)} & .60 {\scriptsize (-10.6\%)} & .41 {\scriptsize (-0.6\%)} & .22 {\scriptsize (+8.6\%)} & .53 {\scriptsize (-0.1\%)} & .50 {\scriptsize (+1.6\%)} \\
 & HyDE Optimized & .65 {\scriptsize (+21.1\%)} & .78 {\scriptsize (+32.0\%)} & .75 {\scriptsize (+11.8\%)} & .46 {\scriptsize (+11.2\%)} & .24 {\scriptsize (+19.5\%)} & .59 {\scriptsize (+10.6\%)} & .58 {\scriptsize (+17.8\%)} \\
\cline{1-9}
\multirow[t]{5}{*}{\qwenlarge{}} & Original & .61 & .86 & .77 & .45 & .24 & .68 & .60 \\
 & Rerank-only & .68 {\scriptsize (+12.5\%)} & .87 {\scriptsize (+0.1\%)} & .78 {\scriptsize (+2.0\%)} & .47 {\scriptsize (+5.5\%)} & .24 {\scriptsize (+0.9\%)} & .68 {\scriptsize (+0.1\%)} & .62 {\scriptsize (+3.3\%)} \\
 & Optimized & .70 {\scriptsize (+15.0\%)} & .88 {\scriptsize (+2.1\%)} & .83 {\scriptsize (+8.1\%)} & .49 {\scriptsize (+8.7\%)} & .25 {\scriptsize (+7.4\%)} & .68 {\scriptsize (+0.5\%)} & .64 {\scriptsize (+6.4\%)} \\
 & HyDE & .58 {\scriptsize (-3.6\%)} & .84 {\scriptsize (-2.8\%)} & .71 {\scriptsize (-7.3\%)} & .44 {\scriptsize (-0.8\%)} & .24 {\scriptsize (+0.9\%)} & .71 {\scriptsize (+3.6\%)} & .59 {\scriptsize (-2.2\%)} \\
 & HyDE Optimized & .68 {\scriptsize (+12.7\%)} & .87 {\scriptsize (+0.7\%)} & .81 {\scriptsize (+4.8\%)} & .48 {\scriptsize (+7.5\%)} & .25 {\scriptsize (+4.7\%)} & .67 {\scriptsize (-1.1\%)} & .63 {\scriptsize (+4.4\%)} \\
\cline{1-9}
\multirow[t]{5}{*}{\nemotron{}} & Original & .54 & .80 & .59 & .45 & .27 & .61 & .54 \\
 & Rerank-only & .65 {\scriptsize (+20.4\%)} & .80 {\scriptsize (+0.5\%)} & .67 {\scriptsize (+13.8\%)} & .46 {\scriptsize (+4.0\%)} & .26 {\scriptsize (-4.5\%)} & .61 {\scriptsize (+0.3\%)} & .58 {\scriptsize (+6.2\%)} \\
 & Optimized & .67 {\scriptsize (+23.0\%)} & .85 {\scriptsize (+6.5\%)} & .74 {\scriptsize (+25.7\%)} & .47 {\scriptsize (+5.1\%)} & .27 {\scriptsize (+0.2\%)} & .63 {\scriptsize (+4.1\%)} & .60 {\scriptsize (+11.5\%)} \\
 & HyDE & .51 {\scriptsize (-5.7\%)} & .74 {\scriptsize (-7.5\%)} & .55 {\scriptsize (-5.5\%)} & .42 {\scriptsize (-6.3\%)} & .24 {\scriptsize (-10.3\%)} & .61 {\scriptsize (+1.1\%)} & .51 {\scriptsize (-5.3\%)} \\
 & HyDE Optimized & .65 {\scriptsize (+20.1\%)} & .82 {\scriptsize (+3.0\%)} & .72 {\scriptsize (+23.8\%)} & .47 {\scriptsize (+5.4\%)} & .26 {\scriptsize (-2.2\%)} & .65 {\scriptsize (+6.7\%)} & .60 {\scriptsize (+10.2\%)} \\
\cline{1-9}
\multirow[t]{5}{*}{\linq{}} & Original & .51 & .76 & .67 & .43 & .24 & .59 & .53 \\
 & Rerank-only & .59 {\scriptsize (+16.8\%)} & .76 {\scriptsize (+0.6\%)} & .73 {\scriptsize (+8.1\%)} & .46 {\scriptsize (+7.0\%)} & .24 {\scriptsize (+0.4\%)} & .60 {\scriptsize (+0.3\%)} & .56 {\scriptsize (+5.5\%)} \\
 & Optimized & .51 {\scriptsize (+1.2\%)} & .76 {\scriptsize (+0.2\%)} & .68 {\scriptsize (+0.6\%)} & .43 {\scriptsize (+0.5\%)} & .24 {\scriptsize (+0.2\%)} & .59 {\scriptsize (+0.2\%)} & .54 {\scriptsize (+0.5\%)} \\
 & HyDE & .50 {\scriptsize (-2.5\%)} & .71 {\scriptsize (-6.1\%)} & .59 {\scriptsize (-12.4\%)} & .42 {\scriptsize (-1.4\%)} & .23 {\scriptsize (-2.5\%)} & .55 {\scriptsize (-8.0\%)} & .50 {\scriptsize (-6.3\%)} \\
 & HyDE Optimized & .50 {\scriptsize (-1.7\%)} & .71 {\scriptsize (-5.8\%)} & .59 {\scriptsize (-11.7\%)} & .42 {\scriptsize (-1.0\%)} & .23 {\scriptsize (-2.0\%)} & .55 {\scriptsize (-7.7\%)} & .50 {\scriptsize (-5.8\%)} \\
\cline{1-9}
\multirow[t]{5}{*}{\efivemistral{}} & Original & .48 & .69 & .63 & .41 & .21 & .55 & .50 \\
 & Rerank-only & .55 {\scriptsize (+15.1\%)} & .70 {\scriptsize (+0.9\%)} & .69 {\scriptsize (+10.4\%)} & .44 {\scriptsize (+7.3\%)} & .22 {\scriptsize (+5.1\%)} & .55 {\scriptsize (+0.7\%)} & .53 {\scriptsize (+6.3\%)} \\
 & Optimized & .60 {\scriptsize (+24.5\%)} & .78 {\scriptsize (+12.8\%)} & .77 {\scriptsize (+23.3\%)} & .46 {\scriptsize (+11.4\%)} & .24 {\scriptsize (+14.1\%)} & .58 {\scriptsize (+5.6\%)} & .57 {\scriptsize (+15.5\%)} \\
 & HyDE & .46 {\scriptsize (-3.1\%)} & .61 {\scriptsize (-11.7\%)} & .56 {\scriptsize (-11.4\%)} & .41 {\scriptsize (-0.6\%)} & .21 {\scriptsize (-1.1\%)} & .44 {\scriptsize (-19.0\%)} & .45 {\scriptsize (-9.3\%)} \\
 & HyDE Optimized & .62 {\scriptsize (+28.4\%)} & .76 {\scriptsize (+8.9\%)} & .74 {\scriptsize (+18.7\%)} & .46 {\scriptsize (+12.3\%)} & .24 {\scriptsize (+14.7\%)} & .57 {\scriptsize (+4.0\%)} & .56 {\scriptsize (+14.1\%)} \\
\cline{1-9}
\bottomrule
\end{tabular}
}
\end{table*}

\subsection{LLM Feedback Scores} \label{app:scores}
To obtain LLM feedback scores for a
specific document, we send the query-document pair along with a brief instruction that asks the LLM to judge whether this pair is a match. The task is framed to the LLM as a yes/no question: \textit{Is this document relevant for this query?}. For calculating the score, we extract the token log-probabilities for ``yes'' and ``no'' tokens (across multiple token realizations in terms of capitalization, punctuation etc.). The feedback score  $[0.0-1.0]$ is the sum of probabilities for \textit{yes} divided by the sum of probabilities for \textit{yes} and \textit{no}.

For classification tasks, we include in the prompt a brief task instruction (the same one that is used for the embedding model query instructions). The full LLM prompt template is provided below.

\begin{tcolorbox}[title=LLM Feedback Prompt,fontupper=\linespread{1.1}\fontfamily{lmr}\selectfont]
Here is a user query and a document. Is this document relevant for this query? Judge the relevance according to the following task: \textit{[task-instruction]}. Answer with only yes/no, without any preceding tokens. Query: \textit{[query]}

Document: \textit{[document]}

Relevant? (Yes/No):
\end{tcolorbox}

\section{Analysis}

\subsection{Visualizing the Query Refinement Through Dimensionality Reduction} \label{app:2d}

\begin{figure*}[t]
\centering
\subfloat[\realscholar{}, \qwenlarge{}]{\includegraphics[width=.48\columnwidth]{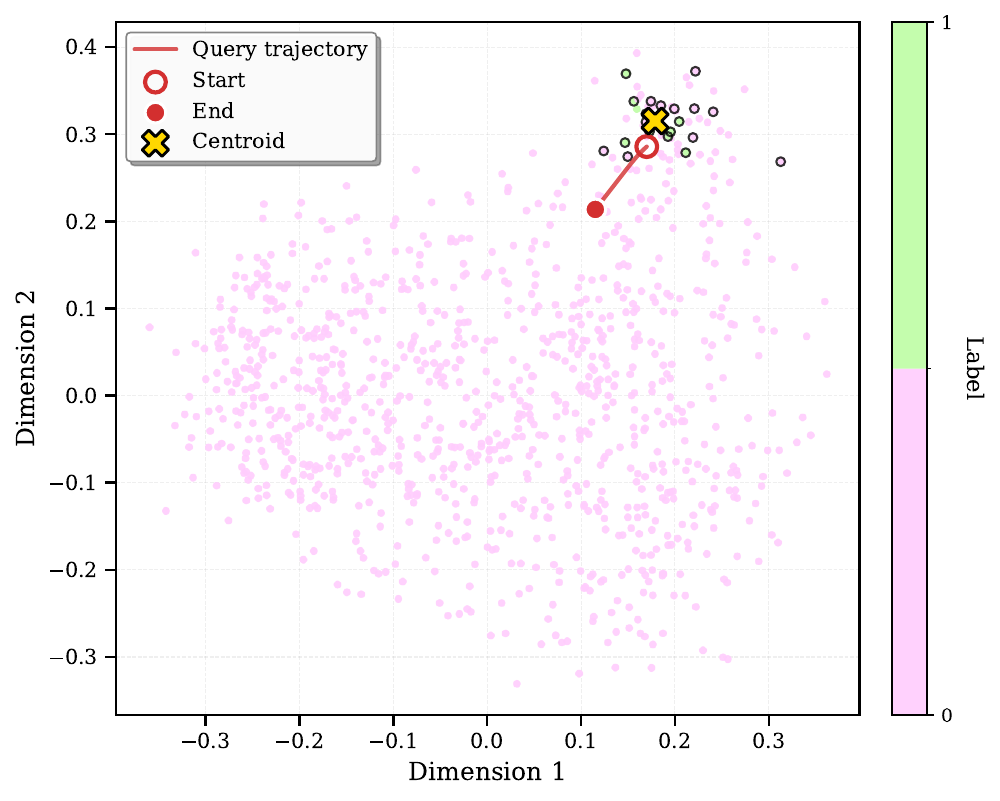}}
\subfloat[\followir{}, \nemotron{}]{\includegraphics[width=.48\columnwidth]{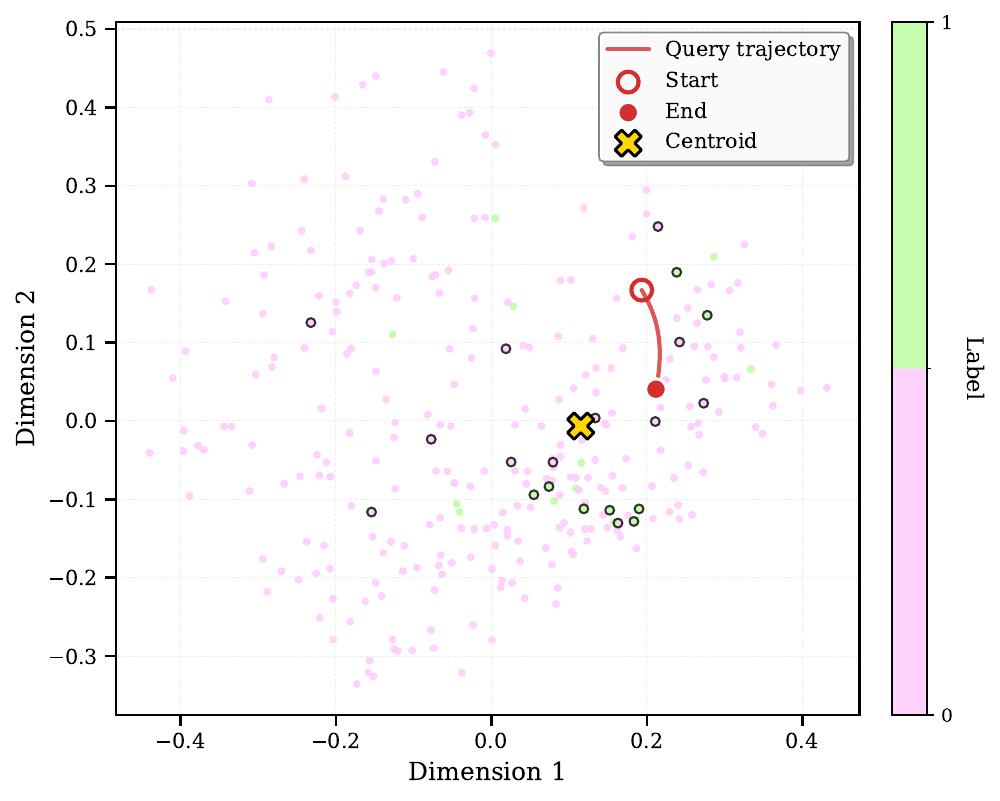}}

\subfloat[\clinc{}, \qwensmall{}]{\includegraphics[width=.48\columnwidth]{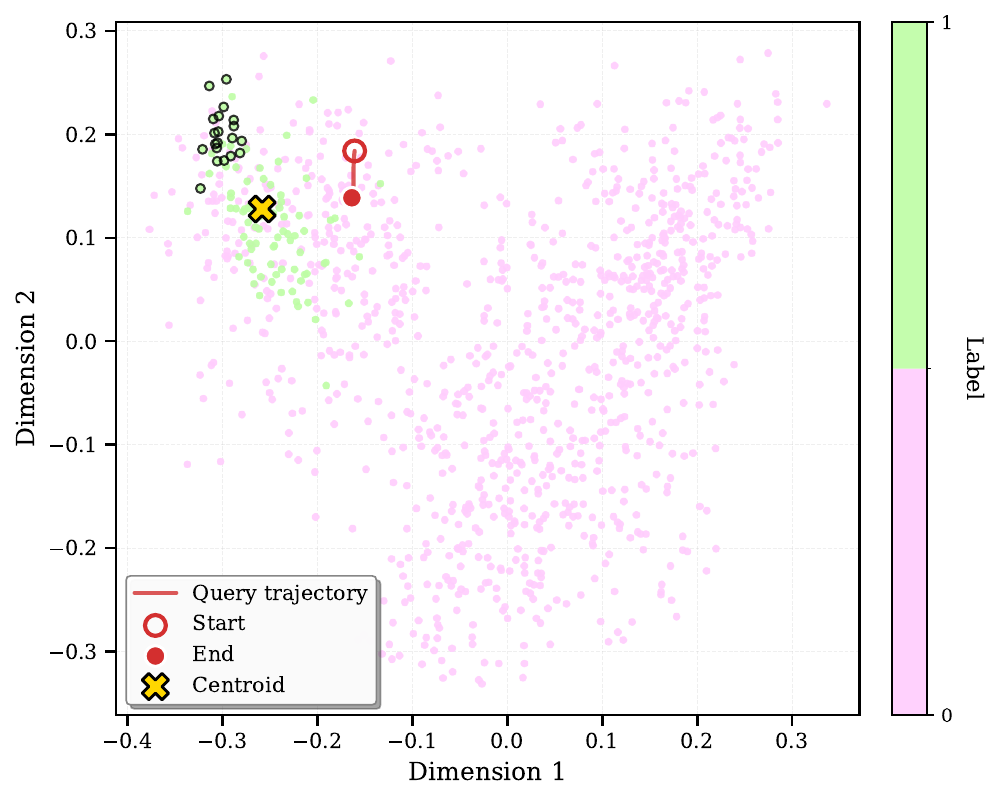}}
\subfloat[\kpm{}, \efivemistral{}]{\includegraphics[width=.48\columnwidth]{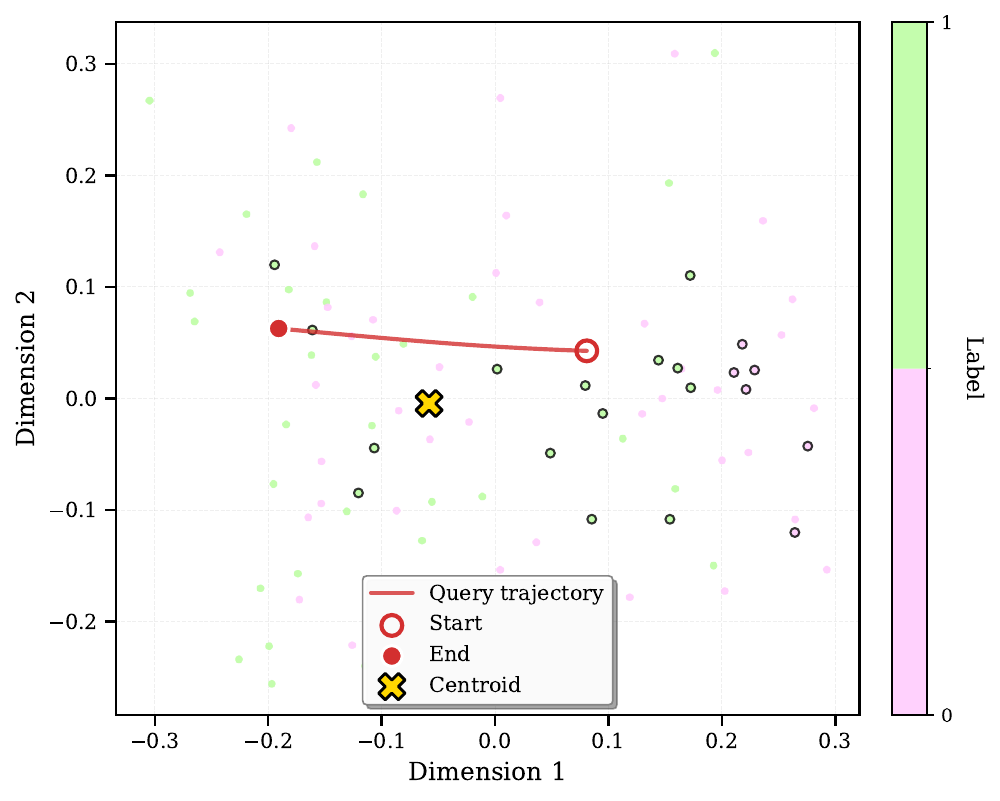}}

\caption{2D PCA projection for example queries. The plots depict the trajectory of the query embedding during the $T=100$ gradient steps of test-time optimization, shown here in the context of the corpus embedding space. Each dot marks a document, and the colors denote the gold label of the query-document pairs. Circled dots are the documents in $\mathcal{D}_K(q)$, which received LLM feedback scores and thus guided the optimization.}
\label{fig:pca_app}
\end{figure*}

Aiming to better understand the underlying mechanism of our method, we apply dimensionality reduction, specifically PCA, to visualize three key components: the corpus, the query's refinement trajectory, and the training data.  
This visualization reveals how queries evolve during refinement, how this evolution relates to the spatial organization of positive and negative examples, and how training samples influence the process. Some examples are shown in Figure~\ref{fig:pca_app}.

Our observations reveal the following insights:

\begin{itemize}
    \item \textbf{Refinement does not necessarily move the query embedding toward the relevant documents centroid.}  
    While we may expect the query to move towards the center of mass of positive documents, this is not always the case (see Figure~\ref{fig:pca_app} d). To quantify this, we also measure the average cosine similarity between the refined query embedding and the centroid, and find that it is comparable to that of the initial query embedding.

    \item \textbf{Refinement brings the query to a better vantage point.}  
    Rather than collapsing toward a central point, the refinement process can relocate the query to a more discriminative region of the embedding space, one that better separates positive examples from negative ones (see Figure~\ref{fig:pca_example}). This improved spatial separation aligns with the empirical performance gains observed in retrieval.

    \item \textbf{The identity of the samples used for feedback has a strong influence.}
    The samples appear to have a substantial impact on the results. Both positive and negative examples contribute to refining the queries and moving them closer to a clear separation. Our observations suggest that achieving an appropriate balance between positive and negative samples may further improve performance.

    Examples for that include Figure~\ref{fig:pca_app} a, where positives and negatives were all in a specific area, which makes the separation goal harder. In Figure~\ref{fig:pca_app} d, it appears that the fact that the negative feedback samples are concentrated in a specific area (right) led to a refinement vector in the opposite direction.

\end{itemize}

\begin{table*}
\centering
\caption{Mean average precision (MAP) results across queries.} \label{tab:results_app_full}

\resizebox{\textwidth}{!}{%

\begin{tabular}{lllllllll}
\toprule
 &  & \realscholar{} & \clinc{} & \kpm{} & \followir{} & \nfcorpus{} & \banking{} & Average \\
Embedding Model & Setting &  &  &  &  &  &  &  \\
\midrule
\multirow[t]{3}{*}{\qwensmall{}} & Original & .54 & .59 & .67 & .41 & .20 & .53 & .49 \\
 & Rerank-only & .62 {\scriptsize (+15.5\%)} & .60 {\scriptsize (+1.3\%)} & .73 {\scriptsize (+8.0\%)} & .45 {\scriptsize (+8.2\%)} & .21 {\scriptsize (+8.0\%)} & .53 {\scriptsize (+0.2\%)} & .52 {\scriptsize (+6.7\%)} \\
 & Optimized & .65 {\scriptsize (+20.7\%)} & .74 {\scriptsize (+25.5\%)} & .79 {\scriptsize (+17.2\%)} & .46 {\scriptsize (+11.3\%)} & .23 {\scriptsize (+16.6\%)} & .60 {\scriptsize (+12.4\%)} & .58 {\scriptsize (+17.8\%)} \\
\cline{1-9}
\multirow[t]{3}{*}{\qwenlarge{}} & Original & .61 & .86 & .77 & .45 & .24 & .68 & .60 \\
 & Rerank-only & .68 {\scriptsize (+12.5\%)} & .87 {\scriptsize (+0.1\%)} & .78 {\scriptsize (+2.0\%)} & .47 {\scriptsize (+5.5\%)} & .24 {\scriptsize (+0.9\%)} & .68 {\scriptsize (+0.1\%)} & .62 {\scriptsize (+3.3\%)} \\
 & Optimized & .70 {\scriptsize (+15.0\%)} & .88 {\scriptsize (+2.1\%)} & .83 {\scriptsize (+8.1\%)} & .49 {\scriptsize (+8.7\%)} & .25 {\scriptsize (+7.4\%)} & .68 {\scriptsize (+0.5\%)} & .64 {\scriptsize (+6.4\%)} \\
\cline{1-9}
\multirow[t]{3}{*}{\nemotron{}} & Original & .54 & .80 & .59 & .45 & .27 & .61 & .54 \\
 & Rerank-only & .65 {\scriptsize (+20.4\%)} & .80 {\scriptsize (+0.5\%)} & .67 {\scriptsize (+13.8\%)} & .46 {\scriptsize (+4.0\%)} & .26 {\scriptsize (-4.5\%)} & .61 {\scriptsize (+0.3\%)} & .58 {\scriptsize (+6.2\%)} \\
 & Optimized & .67 {\scriptsize (+23.0\%)} & .85 {\scriptsize (+6.5\%)} & .74 {\scriptsize (+25.7\%)} & .47 {\scriptsize (+5.1\%)} & .27 {\scriptsize (+0.2\%)} & .63 {\scriptsize (+4.1\%)} & .60 {\scriptsize (+11.5\%)} \\
\cline{1-9}
\multirow[t]{3}{*}{\linq{}} & Original & .51 & .76 & .67 & .43 & .24 & .59 & .53 \\
 & Rerank-only & .59 {\scriptsize (+16.8\%)} & .76 {\scriptsize (+0.6\%)} & .73 {\scriptsize (+8.1\%)} & .46 {\scriptsize (+7.0\%)} & .24 {\scriptsize (+0.4\%)} & .60 {\scriptsize (+0.3\%)} & .56 {\scriptsize (+5.5\%)} \\
 & Optimized & .51 {\scriptsize (+1.2\%)} & .76 {\scriptsize (+0.2\%)} & .68 {\scriptsize (+0.6\%)} & .43 {\scriptsize (+0.5\%)} & .24 {\scriptsize (+0.2\%)} & .59 {\scriptsize (+0.2\%)} & .54 {\scriptsize (+0.5\%)} \\
\cline{1-9}
\multirow[t]{3}{*}{\efivemistral{}} & Original & .48 & .69 & .63 & .41 & .21 & .55 & .50 \\
 & Rerank-only & .55 {\scriptsize (+15.1\%)} & .70 {\scriptsize (+0.9\%)} & .69 {\scriptsize (+10.4\%)} & .44 {\scriptsize (+7.3\%)} & .22 {\scriptsize (+5.1\%)} & .55 {\scriptsize (+0.7\%)} & .53 {\scriptsize (+6.3\%)} \\
 & Optimized & .60 {\scriptsize (+24.5\%)} & .78 {\scriptsize (+12.8\%)} & .77 {\scriptsize (+23.3\%)} & .46 {\scriptsize (+11.4\%)} & .24 {\scriptsize (+14.1\%)} & .58 {\scriptsize (+5.6\%)} & .57 {\scriptsize (+15.5\%)} \\
\cline{1-9}
\bottomrule
\end{tabular}

}

\end{table*}

\begin{table*}
\centering
\caption{Statistical significance of the ranking improvements. The table presents the $p$-values for paired t-tests (one-sided) of the average precision (AP) results, comparing per-query AP values before and after LLM-guided query refinement. $p$-values are corrected for multiple comparisons using the Benjamini-Hochberg FDR method. N.S.: non-significant ($p\geq0.05$).} \label{tab:pvalues}

\resizebox{\textwidth}{!}{%

\begin{tabular}{llllllll}
\toprule
 &  & \realscholar{} & \clinc{} & \kpm{} & \followir{} & \nfcorpus{} & \banking{} \\
 \midrule
\multirow[t]{5}{*}{Original vs. Optimized} & \qwensmall{} & $7.6\times10^{-7}$ & $2.9\times10^{-22}$ & $4.0\times10^{-30}$ & $2.0\times10^{-13}$ & $4.8\times10^{-10}$ & $1.8\times10^{-7}$ \\
 & \qwenlarge{} & $4.3\times10^{-4}$ & $5.8\times10^{-3}$ & $3.3\times10^{-11}$ & $3.4\times10^{-6}$ & $6.3\times10^{-3}$ & N.S. \\
 & \nemotron{} & $1.1\times10^{-6}$ & $2.5\times10^{-5}$ & $4.0\times10^{-36}$ & $5.8\times10^{-3}$ & N.S. & N.S. \\
 & \linq{} & $3.6\times10^{-2}$ & $1.8\times10^{-9}$ & $1.1\times10^{-8}$ & $2.3\times10^{-11}$ & $2.5\times10^{-5}$ & $1.8\times10^{-6}$ \\
 & \efivemistral{} & $1.4\times10^{-7}$ & $2.4\times10^{-8}$ & $1.5\times10^{-31}$ & $3.8\times10^{-7}$ & $2.3\times10^{-6}$ & N.S. \\
\cline{1-8}
\multirow[t]{5}{*}{HyDE vs. Optimized HyDE} & \qwensmall{} & $2.8\times10^{-10}$ & $2.2\times10^{-12}$ & $1.2\times10^{-37}$ & $9.1\times10^{-14}$ & $1.6\times10^{-7}$ & $4.4\times10^{-5}$ \\
 & \qwenlarge{} & $1.1\times10^{-5}$ & $5.9\times10^{-4}$ & $2.5\times10^{-20}$ & $4.5\times10^{-6}$ & N.S. & N.S. \\
 & \nemotron{} & $4.9\times10^{-7}$ & $7.8\times10^{-8}$ & $4.7\times10^{-41}$ & $7.8\times10^{-8}$ & $1.5\times10^{-3}$ & $3.2\times10^{-2}$ \\
 & \linq{} & $1.8\times10^{-4}$ & $6.6\times10^{-13}$ & $1.0\times10^{-18}$ & $1.9\times10^{-8}$ & $1.1\times10^{-2}$ & $3.4\times10^{-10}$ \\
 & \efivemistral{} & $1.3\times10^{-6}$ & $2.1\times10^{-14}$ & $3.4\times10^{-42}$ & $1.1\times10^{-8}$ & $3.6\times10^{-7}$ & $1.4\times10^{-8}$ \\
\cline{1-8}
\multirow[t]{5}{*}{Rerank-only vs. Optimized} & \qwensmall{} & $5.7\times10^{-3}$ & $4.8\times10^{-22}$ & $2.5\times10^{-25}$ & $2.3\times10^{-2}$ & $1.6\times10^{-7}$ & $2.1\times10^{-7}$ \\
 & \qwenlarge{} & N.S. & $1.1\times10^{-2}$ & $5.2\times10^{-12}$ & $2.4\times10^{-2}$ & $3.7\times10^{-3}$ & N.S. \\
 & \nemotron{} & N.S. & $1.0\times10^{-4}$ & $8.1\times10^{-18}$ & N.S. & $7.3\times10^{-3}$ & N.S. \\
 & \linq{} & N.S. & N.S. & N.S. & N.S. & N.S. & N.S. \\
 & \efivemistral{} & $2.5\times10^{-3}$ & $9.9\times10^{-8}$ & $1.4\times10^{-21}$ & $1.7\times10^{-2}$ & $2.0\times10^{-4}$ & N.S. \\
\cline{1-8}
\bottomrule
\end{tabular}

}

\end{table*}

\begin{table*}
\centering
\caption{Mean average precision (MAP) results across queries, comparing the use of different LLM teachers for query optimization. Results shown are with \textit{\qwensmall{}} as the embedding model.} \label{tab:llms}

\resizebox{\textwidth}{!}{%

\begin{tabular}{lllllll}
\toprule
 & \clinc{} & \followir{} & \kpm{} & \realscholar{} & Average \\
\midrule
Original & .589 & .412 & .672 & .541 & .553 \\
\midrule
Rerank-only (\textit{\gptfour{}}) & .595 & .433 & .693 & .601 & .580 \\
Optimized (\textit{\gptfour{}}) & .745 & .461 & .781 & .655 & .661 \\
\midrule
Rerank-only (\textit{\qwenllm{}}) & .598 & .453 & .738 & .623 & .603 \\
Optimized (\textit{\qwenllm{}}) & .749 & .464 & .789 & .658 & .665 \\
\midrule
Rerank-only (\textit{\deepseek{}}) & .597 & .441 & .712 & .616 & .592 \\
Optimized (\textit{\deepseek{}}) & .740 & .456 & .719 & .634 & .637 \\
\midrule
Rerank-only (\textit{\mistralsmall{}}) & .597 & .445 & .726 & .625 & .598 \\
Optimized (\textit{\mistralsmall{}}) & .739 & .458 & .788 & .653 & .659 \\
\bottomrule
\end{tabular}
}

\end{table*}

\begin{table*}
\centering
\caption{Correlation between feedback quality and performance gains. The table depicts Pearson correlations between the per-query AUC of the LLM feedback scores (with respect to the ground-truth relevance labels), and the percentage gain in the average precision (AP) for the corresponding query as a result of query refinement.} \label{tab:perf_correlations}


\begin{tabular}{lllllll}
\toprule
Dataset      & LLM Teacher                                      & Pearson r & p‑value   \\
\midrule
\clinc{}     & \gptfour{}                   & $0.31$      & $2.3\times10^{-3}$   \\
\clinc{}     & \qwenllm{}                   & $0.24$      & $2.2\times10^{-2}$   \\
\clinc{}     & \deepseek{}                    & $0.26$      & $1.9\times10^{-2}$   \\
\clinc{}     & \mistralsmall{} & $0.33$      & $1.4\times10^{-3}$   \\
\followir{}     & \gptfour{}                   & $0.39$      & $1.8\times10^{-9}$   \\
\followir{}     & \qwenllm{}                   & $0.35$      & $4.6\times10^{-7}$   \\
\followir{}     & \deepseek{}                    & $0.34$      & $7.4\times10^{-7}$   \\
\followir{}     & \mistralsmall{} & $0.36$      & $6.4\times10^{-7}$   \\
\kpm{}          & \gptfour{}                   & $0.19$      & $5.5\times10^{-2}$ (N.S.)  \\
\kpm{}          & \qwenllm{}                   & $0.19$      & $5.5\times10^{-2}$ (N.S.)  \\
\kpm{}          & \deepseek{}                    & $0.19$      & $5.1\times10^{-2}$  (N.S.) \\
\kpm{}          & \mistralsmall{} & $0.16$      & 0.12 (N.S.)   \\
\realscholar{} & \gptfour{}               & $0.25$      & $9.7\times10^{-2}$ (N.S.)  \\
\realscholar{} & \qwenllm{}               & $0.25$      & $9.7\times10^{-2}$ (N.S.)  \\
\realscholar{} & \deepseek{}                & $0.25$      & $9.7\times10^{-2}$ (N.S.)  \\
\realscholar{} & \mistralsmall{} & $0.30$ & $3.8\times10^{-2}$   \\
\bottomrule
\end{tabular}

\end{table*}



\end{document}